%% file: main.tex
\documentclass[times,preprint,12pt]{elsarticle}

\usepackage[acronym,indexonlyfirst,nomain]{glossaries}
\input{acronyms}
\usepackage{hyperref}

\usepackage{diagbox}
\usepackage{rotating}
\usepackage{amssymb}
\usepackage{booktabs, array} 
\usepackage{subcaption,booktabs}
\usepackage{xcolor,colortbl}
\usepackage{amssymb}
\usepackage{amsthm}
\usepackage{amsmath}

\usepackage[a4paper, margin=3cm,lmargin=3cm,rmargin=3cm, hmargin=3cm,vmargin=3cm]{geometry}
\journal{Engineering Applications of Artificial Intelligence}

\usepackage{nomencl}
\makenomenclature

\usepackage{etoolbox}
\renewcommand\nomgroup[1]{%
  \item[\bfseries
  \ifstrequal{#1}{A}{Abbreviations}{%
  \ifstrequal{#1}{N}{Notations}{%
  \ifstrequal{#1}{V}{Variables}{}}}%
]}
\begin{document}

\begin{frontmatter}


% \centering
% \includegraphics[scale=0.81]{images/FC_abstract0511.JPG}

% \begin{highlights}
% \item A learning methodology is investigated to address the data scarcity issue in fatigue damage prognostics problems.
% \item The Self-Supervised Learning paradigm is proposed to improve the learning performance of deep learning models.
% \item Self-supervised pre-trained models are compared with non-pre-trained models in downstream remaining useful life (RUL) prediction task.
% \item A synthetic dataset is used, composed of multivariate run-to-failure time series strain data for structures subject to fatigue crack propagation.
% \end{highlights}

\title{Self-Supervised Learning for Data Scarcity in a Fatigue  \\Damage Prognostic Problem}


\author[1,2]{Anass Akrim\corref{cor1}}
\ead{anass.akrim@gmail.com}
\author[1]{Christian Gogu} 
\author[2]{Rob Vingerhoeds} 
\author[1,2]{Michel Salaün} 
\cortext[cor1]{Corresponding author} 
\address[1]{Institut Clément Ader (UMR CNRS 5312) INSA/UPS/ISAE/Mines Albi, 
\\
Université de Toulouse, 3 rue Caroline Aigle, 31400 Toulouse, France\\}
\address[2]{ISAE-SUPAERO, Université de Toulouse, 10 Avenue Edouard Belin,\\
31400 Toulouse, France}

\date{December 14, 2022}
% \today}

% \author{}

% \affiliation{organization={},%Department and Organization

\begin{abstract}
With the increasing availability of data for Prognostics and Health Management (PHM), Deep Learning (DL) techniques are now the subject of considerable attention for this application, often achieving more accurate \gls{rul} predictions. However, one of the major challenges for DL techniques resides in the difficulty of obtaining large amounts of labelled data on industrial systems. To overcome this lack of labelled data, an emerging learning technique is considered in our work: Self-Supervised Learning, a sub-category of unsupervised learning approaches. 
This paper aims to investigate whether pre-training DL models in a self-supervised way on unlabelled sensors data can be useful for RUL estimation with only Few-Shots Learning, \textit{i.e.} with scarce labelled data. In this research, a fatigue damage prognostics problem is addressed, through the estimation of the \gls{rul} of aluminum alloy panels (typical of aerospace structures) subject to fatigue cracks from strain gauge data. Synthetic datasets composed of strain data are used allowing to extensively investigate the influence of the dataset size on the predictive performance. Results show that the self-supervised pre-trained models are able to significantly outperform the non-pre-trained models in downstream RUL prediction task, and with less computational expense, showing promising results in prognostic tasks when only limited labelled data is available. 

\end{abstract}


\end{frontmatter}


\begin{small}
\nomenclature[A]{\(AE\)}{Autoencoder}
\nomenclature[A]{\(AR\)}{Autoregressive}
\nomenclature[A]{\(DGN\)}{Deep Gated Recurrent Unit Network}
\nomenclature[A]{\(DL\)}{Deep Learning}
\nomenclature[A]{\(GRU\)}{Gated Recurrent Unit}
\nomenclature[A]{\(LSTM\)}{Long Short-Term Memory}
\nomenclature[A]{\(MAPE\)}{Mean Absolute Percentage Error}
\nomenclature[A]{\(ML\)}{Machine Learning}
\nomenclature[A]{\(MSE\)}{Mean Squared Error}
\nomenclature[A]{\(MSPA\)}{Multi-Steps Prediction Autoregressive}
\nomenclature[A]{\(PHM\)}{Prognostics and Health Management}
\nomenclature[A]{\(RNN\)}{Recurrent Neural Networks}
\nomenclature[A]{\(RUL\)}{Remaining Useful Life}
\nomenclature[A]{\(SSL\)}{Self-Supervised Learning}

\nomenclature[N]{\(D_U\)}{Unlabelled Dataset}
\nomenclature[N]{\(D_L\)}{Labelled Dataset}
\nomenclature[N]{\(X^U\)}{Unlabelled input signal}
\nomenclature[N]{\(X^L/y^L\)}{Labelled input signal/Corresponding RUL label}
\nomenclature[N]{\(T_f\)}{Time of failure}

% \nomenclature[N]{\(y^L\)}{}

% \nomenclature[N]{\(\mathbb{C}\)}{Complex numbers}

\nomenclature[V]{\(d\)}{Ratio of the total lifetime of a sequence}
\nomenclature[V]{\(h\)}{Length of the sliding window}
\nomenclature[V]{\(n_g\)}{Number of sensor time series}

\nomenclature[V]{\(N_U\)}{Number of unlabelled structures}
\nomenclature[V]{\(n_U\)}{Number of unlabelled samples}
\nomenclature[V]{\(N_U^{Train}\)}{Number of unlabelled structures for training}
\nomenclature[V]{\(n_U^{Train}\)}{Number of unlabelled samples for training}
% hey
\nomenclature[V]{\(N_L\)}{Number of labelled structures}
\nomenclature[V]{\(n_L\)}{Number of labelled samples}
\nomenclature[V]{\(N_L^{Train}\)}{Number of labelled structures for training}
% \nomenclature[V]{\(n_L^{Train}\)}{Number of labelled samples for training}

\nomenclature[V]{\(N_L^{Test}\)}{Number of labelled structures for testing}

% \nomenclature[V]{\(N_U\)}{Number of unlabelled structures for training}

\printnomenclature
\end{small}


\section{Introduction}

Prognostics and Health Management (PHM) is a reseach domain addressing failure mechanisms of real systems in order to better manage the use of information on equipment operating conditions \cite{shao2013development}. Its implementation can improve the efficiency of maintenance support \cite{mao2010research}, optimize the maintenance plan and therewith equipment availability \cite{atamuradov2017prognostics}, help industry to balance safety and economic profit \cite{zhen2011applications}. For many mechanical structures and notably aerospace structures, fatigue damage is one of the major modes of failure. Therefore, fatigue monitoring and prediction of fatigue life in structures, \textit{i.e.} Remaining Useful Life (RUL) estimation, represents one of the major challenges to be solved for paving the way towards predictive structural maintenance. \\

Among the approaches used for PHM, Data-Driven models have gained more and more attention in the PHM community, especially the latest Deep Learning (DL) techniques \cite{tsui2015prognostics}, redefining state-of-the-art performances in a wide range of areas in recent years \cite{LeCun2015}. However, their effectiveness depends on the quantity and quality of available labelled data. Currently, data scarcity represents a scientific bottleneck in many engineering fields (\textit{e.g.} in healthcare \cite{jadon2021covid}, in energy \cite{berthou2019urban}, water and environmental engineering \cite{borzooei2019data, gutierrez2020improving}, etc.), which makes it difficult to apply the latest Machine Learning (ML) methods. Many approaches have been proposed to address data scarcity in these various domains, as recently reviewed by \cite{nandy2022audacity, gorgoglione2020overcoming, bansal2022systematic}. As faults are rare and structures can be replaced before reaching failure, data scarcity is becoming one of the most important challenges in PHM \cite{fink2020potential,theissler2021predictive}. Nevertheless, while labelled data is lacking, the availability of raw sensors data is increasing due to the advancements in sensing technologies. This data is considered as “unlabelled” in the context of prognostics as, for sensor data at a given point in time, the true RUL is unknown and cannot be determined unless the sensor measurements are available all the way to failure. In most engineering applications, this is unattainable, since the parts will be replaced before failure, and this is particularly true for aerospace mechanical structures, thus the majority of sensor data is unlabelled, meaning that no associated RUL is available for it. Exploiting such unlabelled sensors data during training has become a major goal in ML in order to improve learning performance. Therefore, the research question addressed in this paper can be stated as follows: \textit{is it possible to learn meaningful representations from unlabelled data and use it to enhance related supervised predictive tasks on a fatigue damage prognostics problem?} \\

% Nevertheless, while labelled data is lacking, the availability of unlabelled data is increasing due to the advancements in sensing technologies: exploiting unlabelled data during training (\textit{e.g. }raw sensors data from structures replaced before reaching failure) has become a major goal in ML in order to improve learning performance. Therefore, the research question addressed in this paper can be stated as follows: \textit{is it possible to learn meaningful representations from unlabelled data and use it to enhance related supervised predictive tasks on a fatigue damage prognostics problem?} \\

In the Artifical Intelligence (AI) community, a recent learning technique to extract knowledge from unlabelled data was proposed to address the challenge of data scarcity: Self-Supervised Learning (SSL) \cite{jaiswal_survey_2021}, a sub-category of unsupervised learning approaches. SSL has already shown tremendous performances in many AI fields such as in Natural Language Processing (\textit{e.g. }GPT-3 \cite{brown2020language}) or Image Processing \cite{chen2020big}. Nevertheless, the applicability of this approach remains largely unexplored in the engineering fields, a domain in which data scarcity is an increasingly challenging issue \cite{borzooei2019data, zhu2021novel, rocchetta2022robust}. Currently, there is only a limited amount of existing research that focuses on the potential of Self-Supervised Learning for Prognostics \cite{krokotsch_improving_2022, guo2022masked}, and particularly for fatigue damage prognostics problems. \\

In order to address this limitation, this paper aims to investigate whether pre-training DL models in a self-supervised way on unlabelled sensors data on a fatigue damage prognostics problem can be useful for RUL estimation with only Few-Shots Learning, \textit{i.e.} with scarce labelled data. The interest is in estimating the RUL of aluminum alloy panels (typical of aerospace structures) subject to fatigue cracks from strain gauge data. A synthetically generated dataset is used for this purpose, composed of a large unlabelled dataset (i.e. strain gauges data of structures before failure) for pre-training, and a smaller labelled dataset (i.e. strain gauges data of structures until failure) for fine-tuning on the RUL prediction task. The synthetic dataset is based on a framework previously developed by the authors \cite{ICPHM}. \\

\begin{figure}[!htp]
    \centering
    % \floatpagestyle{plain}
    %  \hspace{-2.2cm}
    \includegraphics[width=150mm]{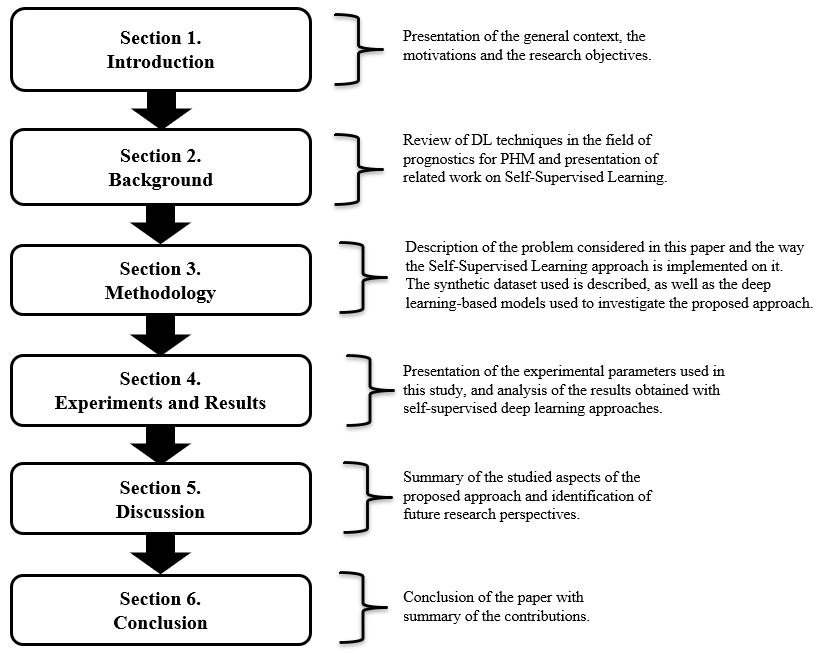}
    \caption{Outline of this paper.}
    \label{fig:outline}
\end{figure}

The remainder of the paper is structured as follows. Section \ref{sec:back} provides a background on the state-of-the-art DL techniques in prognostics for PHM. In Section \ref{sec:approach}, the proposed methodology is presented. Section \ref{sec:exp} presents the experimental settings used in this study for pre-training and fine-tuning phases, and the results obtained with the deep learning-based approaches trained in a self-supervised manner are analyzed. The impact of the size of the data available for pre-training as well as the choice of the pre-text task will be investigated, and different DL models are compared for the self supervised learning task. Section \ref{sec:discuss} summarizes the aspects of the approach considered in this paper and identifies potential future work. Finally, Section \ref{sec:conclusions_outlook} concludes the current research paper and provides future outlooks. Fig. \ref{fig:outline} illustrates the outline of this paper and summarizes the objective of each section.

\section{Background}
\label{sec:back}

In this section, the application of DL techniques in the field of prognostics for PHM and related work on Self-Supervised Learning are presented.

\subsection{Deep Learning in Prognostics for PHM}

As more data becomes available in the engineering domain, there is a recent surge of interest in using Deep Learning in Prognostics and Health Management \cite{MONTEROJIMENEZ2020539, Voulodimos2018}. In prognostics applications, Time Series Forecasting models are most commonly used to predict the RUL of systems or structures, given the format of acquired data in PHM (\textit{e.g. }data collected from sensors, vibration signals, etc.), and the most commonly used algorithms for these tasks are Recurrent Neural Networks (RNN) \cite{hewamalage2021recurrent}. Given the sequential nature of the sensor data in the prognostics field (\textit{e.g. }sensors data), good results have been obtained within the PHM community by using RNNs, such as Standard RNNs, Long Short-Term Memory (LSTM) networks, and Gated Recurrent Unit networks (GRU) \cite{fink2020potential, rana2016gated, baptista2020prognostics}. \\ 

The lack of available labelled data is becoming a major challenge in the application of machine learning to PHM, which, as a field, suffers from a high data acquisition cost compared to other domains in which machine learning has proven useful (\textit{e.g. }Natural Language Processing). In Prognostics tasks, a label can constitute the RUL at each time step of measurements, which is generally difficult to acquire and often can be a time-consuming and expensive investment for experts. However, due to advancements in sensing technologies in engineering fields, the availability of unlabelled data is increasing (\textit{e.g. }raw sensors data of structures replaced before reaching failure). Since failure is not reached when the structures are replaced, the RUL at replacement and at each previous timestep is not known. Thus the sensor data is unlabelled according to the previously introduced definition of a label in the PHM context. Exploiting unlabelled data during training has therefore become a major goal in order to improve learning performance.

\subsection{Self-Supervised Learning}

Similarly to \textit{self-taught learning} as presented in \cite{raina2007self}, Self-Supervised Learning (SSL) consists in learning meaningful and general representations from unlabelled data (during a \textit{pre-training phase}) by solving a so-called \textit{pretext task} without requiring the data to be labelled. These representations are then applicable to a wide range of related supervised tasks (\textit{i.e.} \textit{downstream} \textit{task}) with only few labelled data (\textit{i.e.} ``Few-Shots Learning"). SSL aims to improve predictive performance on the downstream task through the use of unlabelled data, thus avoiding the extensive cost of collecting and annotating large-scale datasets \cite{jing2020self}. This learning paradigm has already proven that it can significantly improve the performance of downstream tasks for many AI applications such as in Natural Language Processing (\textit{e.g. }GPT-3 \cite{brown2020language}) or Image Processing \cite{chen2020big, chen2020generative}. GPT-3 \cite{brown2020language} was one of the largest self-supervised learning systems released by OpenAI in 2019. \\

There are few recent developments that have shown the potential of the SSL paradigm in engineering fields \cite{yengera2018less, endo2022gaitforemer, yu2022self, shurrab2022self}, and several PHM researchers considered this approach to address the data scarcity in fault diagnostics problems, showing promising results \cite{hahn2021self,ding2022self}.
However, to date, there is, to the best of the authors' knowledge, only a limited amount of existing research that focuses on the application of SSL to prognostic problems in PHM, for example for RUL estimation on the NASA C-MAPSS\footnote{NASA C-MAPSS \cite{Saxena2008} is a publicly available dataset of simulated turbofan engines commonly used to benchmark RUL estimation algorithms. The dataset is divided into four subsets (FD001 - FD004) of different operating conditions and possible fault modes.} dataset \cite{yoon2017semi, ellefsen2019remaining, krokotsch_improving_2022, guo2022masked}. 
One of the first to explore this learning paradigm in prognostics in order to deal with the problem of lack of labelled data is Yoon et al. \cite{yoon2017semi}, using a pre-trained variational autoencoder (VAE \cite{kingma2013auto}) that makes use of available unlabelled data to learn a latent space representation in an unsupervised manner; the pre-text task being the minimization of the reconstruction error. The extracted features by the VAE model are then fed as inputs to an RNN model for RUL estimation, trained in a supervised manner by varying the fraction of labelled engines data down to 1\% in order to investigate if the SSL approach can enhance predictive tasks when only a small amount of labelled data is available. Results showed that their proposed method was able to outperform a non pre-trained supervised RNN-model when all the labelled dataset is available as well as in other scenarios when the available labelled data is highly limited with only a small labelled fraction of the training data. The authors in \cite{ellefsen2019remaining} used a Restricted Boltzmann Machine model (RBM) \cite{hinton2009deep} for pre-training on unlabelled dataset with a reconstruction pre-text task, and an LSTM model for RUL prediction. Results showed that this SSL approach could improve the RUL prediction accuracy compared to the purely supervised learning approach (\textit{i.e.} predictive model without the initial pre-training stage), both when the training data is completely labelled and when the labelled training data is reduced. It is worth noting that the methods proposed in \cite{yoon2017semi} and \cite{ellefsen2019remaining} were not presented as SSL approaches, but are considered as such in this paper since the proposed methods follow the same procedure as described earlier. However, Krokotsch et al. \cite{krokotsch_improving_2022} highlighted two shortcomings of these two previous studies: 
\begin{enumerate}
    \item the approaches were evaluated only on one subset of the C-MAPSS dataset out of four in each study, rendering these investigations limited;
    \item pre-training was performed on unlabelled data of engines that contain the point of failure, which should not be the case in real scenarios, since the RUL labels for all the data could be deduced based on the knowledge of the failure time.
\end{enumerate}

To overcome these limitations in \cite{krokotsch_improving_2022}, the investigation was performed over all subsets of the C-MAPSS data set and the unsupervised pre-training phase was performed over truncated time series, assuming that realistic unlabelled data does not contain features near the time of failure (corresponding to sensors data of structures replaced before reaching failure). Results showed that:
\begin{enumerate}
    \item the proposed SSL approach can outperform the supervised baseline that used only the labelled data. Both approaches were trained on only few labelled time series for RUL estimation (\textit{i.e. }Few-Shots learning);
    \item the proposed pre-training model outperformed two competing pre-training models, including AE and RBM using a reconstruction pre-text task (\textit{i.e.} the output $y$ corresponds to an estimation of the input $x$). 
\end{enumerate}

These results suggest that the choice of the pre-training model (or pre-text task) matters. Recently, Guo et al. \cite{guo2022masked} proposed a pre-training method based on masked autoencoders \cite{he2022masked} to perform SSL on the C-MAPSS datasets. Results showed that their pre-trained model outperformed the fully supervised model in RUL estimation. Unfortunately, there are no clear guidelines for selecting the right pre-text task that learns meaningful representations from unlabelled time series data (\textit{e.g. }sensors data) during the pre-training phase. Furthermore, one of the main challenges for extensive investigations on the potential of SSL in PHM resides in the difficulty of having scalable open-source dataset, similar to those available in Natural Language Processing or Image Processing. Thus, despite demonstrating encouraging results, the domain of SSL is still largely unexplored in the prognostics field and is in contrast with the increasing amount of unlabelled data available in industry, having the potential to enable predictive maintenance. 
Table \ref{tab:list} summarizes the applications of self-supervised learning in PHM identified in this paper.\\

\begin{table}[!htp]
    \linespread{1}

    \footnotesize
    \hrulefill
	\centering

	\begin{tabular}{p{2.5cm} p{1cm} p{2.5cm} p{4cm} p{1.5cm}}
	    \hline
		\textbf{Authors} & \textbf{Year} & \textbf{Downstream task} & \textbf{Pre-training tasks} & \textbf{Application}\\
		\hline
		\hline

        Yoon et al. \cite{yoon2017semi} & 2017 & RUL estimation & Reconstruction of the input signal (variational autoencoder).  & Turbofan engines\\[1cm]
        
        Ellefsen et al. \cite{ellefsen2019remaining} & 2019 & RUL estimation & Reconstruction of the input signal (restricted boltzmann machine).  & Turbofan engines\\[1cm]

        Krokotsch et al. \cite{krokotsch_improving_2022} & 2022 & RUL estimation & Reconstruction of the input signal (autoencoder and restricted boltzmann machine);  learn a distance or similarity metric between pairs of data (siamese network).  & Turbofan engines
        \\\\
        % [1cm]

        Guo et al. \cite{guo2022masked} & 2022 & RUL estimation & Reconstruction of the input signal (masked autoencoder).  & Turbofan engines\\[1cm]

        Hahn et al. \cite{hahn2021self} & 2021 & Fault diagnostics & Reconstruction of the input signal (variational autoencoder).  & Milling tools\\[1cm]

		Ding et al. \cite{ding2022self} & 2022 & Fault diagnostics & Contrastive learning (deep convolutional network).  & Bearings\\[0.5cm]

		\hline
		
	\end{tabular}
	\caption{\footnotesize Summary of identified applications of SSL approaches in PHM.}
	\label{tab:list}
\end{table} 
\newpage

\section{Methodology}
\label{sec:approach}

The authors of this paper seek to advance the field of data scarcity in fatigue damage prognostic problems by investigating Deep Self-Supervised Learning on an associated RUL estimation problem. In this section, a description of the dataset involved is provided, followed by a description of the problem considered in this paper and the way the Self-Supevised Learning approach is implemented on it. The deep learning-based models used to investigate the SSL approach are also presented and detailed in this section. Note that data and code for the learning procedure are publicly available on \url{https://github.com/ansak95/DeepSSL}.

\subsection{Data Description}

\begin{figure}[!htp]
\centering
\includegraphics[width=150mm]{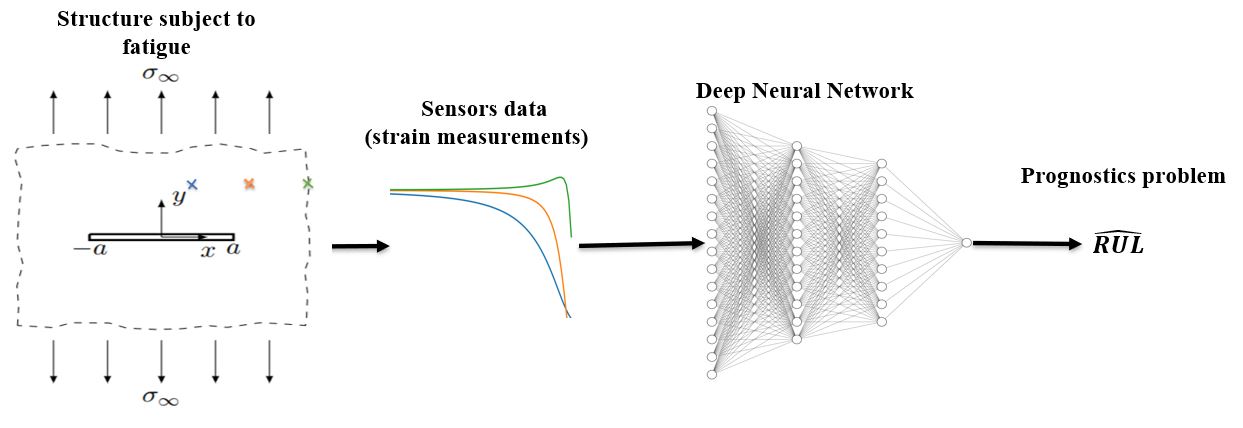}
\caption{Illustration of 3 run-to-failure time series generated (\textit{i.e.} strain data) used as sensors data, and as input for prognostic problems (\textit{e.g.} RUL estimation). }
\label{fig:ill}
\end{figure}

In the current research study, a synthetic dataset for a realistic fatigue damage prognostics problem is generated, based on a framework previously proposed by the authors \cite{ICPHM}. It consists of synthetic multivariate run-to-failure time series data for structures subject to fatigue crack propagation (\textit{e.g.} fuselage panels). Indeed, the proposed framework generates synthetic data sets of mechanical strain data (i.e. virtual strain gauges), by simulating the crack growths in structures based on the Paris-Erdogan model \cite{Paris1963}. Strain data was considered as sensor data since we consider a mechanical fatigue propagation problem and strain data is one of the main, easily measurable, quantities of interest allowing to determine crack propagation. Furthermore, strain gage measurement is a mature technique that can be relatively easily implemented on various kinds of structures. The strain data, or measurement sequences, are obtained until the crack size $a$ reaches the critical crack size $a_{crit}$, considered as the time of failure (necessary to compute the RUL at each time step for example).
Finally, the generated strain data are used as sensors time series data available for prognostics problem such as RUL estimation. This setup can be seen representative of real experiments under fatigue loading where the strain state is monitored at multiple strain gauge positions (blue, orange and green crosses), illustrated in Fig. \ref{fig:ill}. \\

% \newpage

\begin{figure}[!htp]
		\centering
        % \hspace{-2.5cm}
		\includegraphics[width=0.6\textwidth]{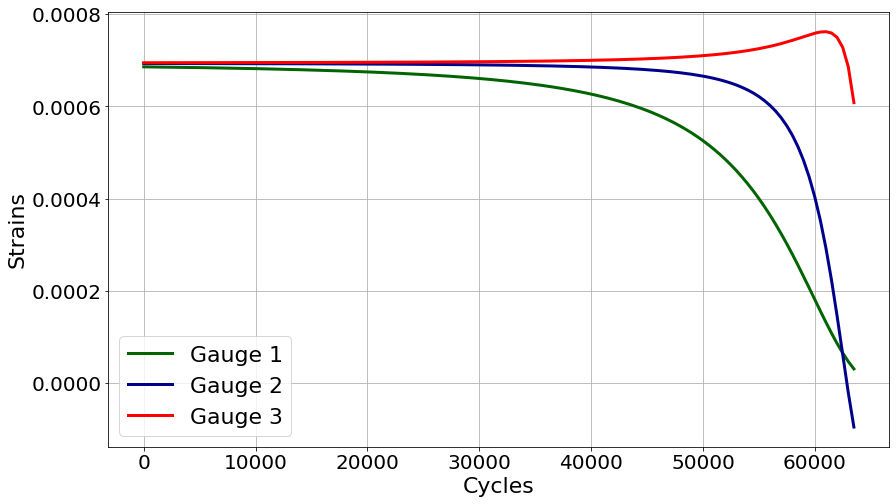}
		\caption{Strain values time series corresponding to a random sensor sample generated reaching failure (\textit{i.e.} a labelled sequence).}
	   \label{fig:fig33}
	\end{figure}

In the current research, the multivariate dataset used contains the variations of the strains at $n_g = 3$ positions in the panel as a function of the number of cycles, where $n_g$ is the number of the time series. More details about the dataset are given in \cite{ICPHM}, and an illustration of a generated sequence (\textit{i.e.} three placed gauges) for a single structure until failure is given in Fig. \ref{fig:fig33}.\\

% \newpage

\begin{figure}[!htp]
		\centering
        % \hspace{-2.5cm}
		\includegraphics[width=0.6\textwidth]{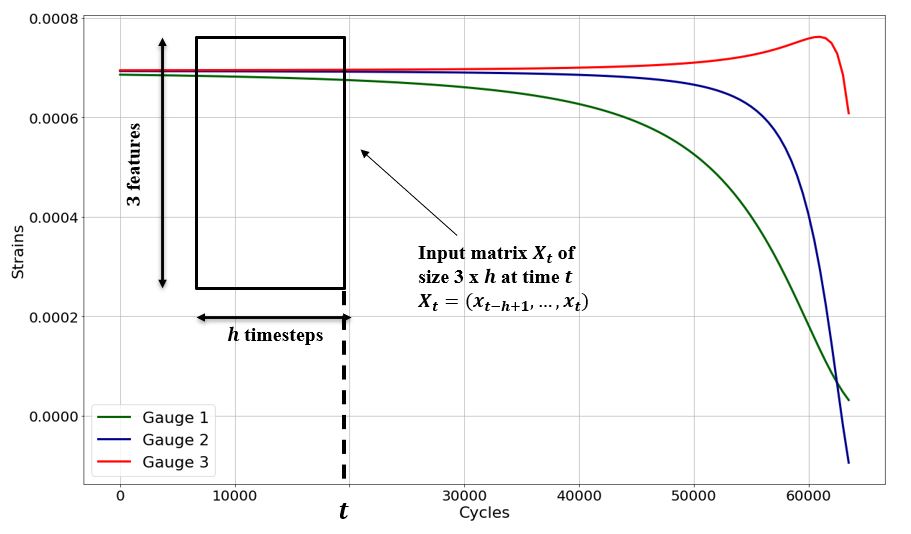}
		\caption{Illustration of the processing of the input data using a ”sliding window” of size $h$ to predict the corresponding output at each timestep $t$.}
	   \label{fig:window}
	\end{figure}

Given the sequential nature of the sensors data, the time series generated are processed sequentially on a \textit{sliding window} approach of size $h$:  at each time-step $t$, the input of the predictive models corresponds to the current and past measurements, such that $ X_t:= (x_{t-h+1}, \dots , x_t) \in \mathbb{R}^{n_g ~ \text{x} ~ h} $ where $ h = 30$ is the length of the sliding window (note that the value of parameter $h$ was set after preliminary experiments). Fig. \ref{fig:window} illustrates the sliding window approach used in this work. \\

\subsection{The proposed Self-Supervised Learning Approach}

The Self-Supervised learning paradigm aims to extract useful features from unlabelled data in a self-supervised manner that can subsequently benefit supervised training on few labelled samples. Hence it is typically composed of:

\begin{enumerate}
    \item \textit{\underline{a pre-training phase}}: a predictive data-driven model is trained on a raw unlabelled dataset in an unsupervised (or \textit{self-supervised}) manner in order to learn abstract features. 
    
    \item \textit{\underline{a fine-tuning phase}}: the pre-trained model is coupled to a non-pre-trained model (\textit{e.g.} for neural networks a linear layer or data-driven model) and then trained on a set of labelled data in a supervised manner.
\end{enumerate}

\begin{figure}[!htp]
	\centering
	\subfloat[][]{\includegraphics[width=0.7\textwidth]{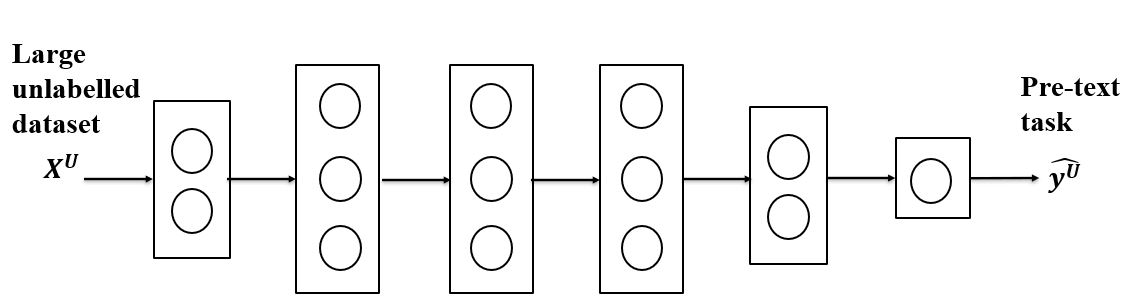}}
	\\
	\subfloat[][]{\includegraphics[width=0.7\textwidth]{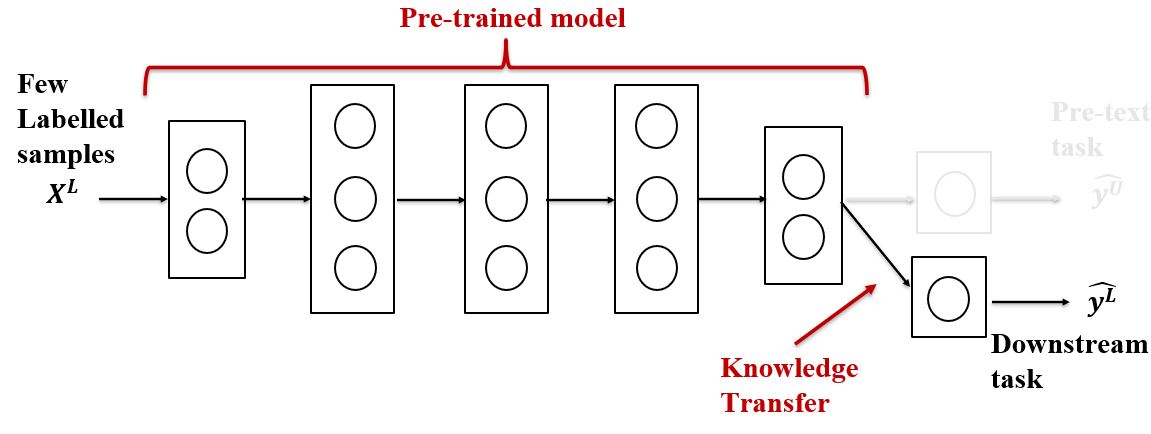}}
	\caption{\footnotesize A schematic view of the Self-Supervised Learning procedure (a): Pre-training phase in a self-supervised way, (b): Fine-tuning phase (supervised training on downstream tasks).}
	\label{fig:SSL}
\end{figure}

The pre-training in SSL is essentially performed with deep learning models. Indeed, the architecture of DL models is in the form of a stack of layers of neurons, and the last layer is used to obtain the final output. Knowledge transfer is typically performed by removing this last layer and replacing it with a new non-trained output linear layer (or a predictive model). The working of SSL can be illustrated in Figure \ref{fig:SSL}. This strategy allows to reuse the learned knowledge in terms of global architecture of the pre-trained network, which works as a \textit{features extractor}, and to exploit it as a starting point for a downstream task (\textit{i.e.} fine-tuning phase). It also provides faster learning time in downstream predictive tasks compared to non-pre-trained models, since it is not necessary to train the pre-trained layers but only the new output linear layer (or predictive model). This aspect will be discussed in Section 4.4.2. Note that some machine learning models are not suitable for pre-training in SSL paradigm, as their architecture is not composed of layers that can be easily extracted and reused for knowledge transfer (\textit{e.g.} Support Vector Machines \cite{cortes1995support}, Random Forests \cite{ho1995random}, Gaussian Processes \cite{rasmussen2003gaussian}). Nevertheless, there are recent developments of these models that can be used in knowledge transfer (\textit{e.g.} Deep Gaussian Processes \cite{damianou2013deep,kandemir2015asymmetric}).

% \newpage

\subsubsection{Problem statement}

To clearly formulate the problem, the synthetic data used in this work is composed of:

\begin{enumerate}
    % \vspace{-0.5cm}
    \item A large set of unlabelled data $D_U = \{X_i^U\}_{i=1}^{n_U}$, where $n_U$ is the number of unlabelled samples, $X_i^U \in \mathbb{R}^{{n_g} ~ \text{x} ~ h}$ the input signal with $n_g$ sensors and $h$ time steps. The unlabelled set $D_U$ refers to strain measurement sequences of structures before reaching failure.
    \item A smaller set of labelled data $D_L = \{(X_i^L,y_i^L)\}_{i=1}^{n_L}$, where $n_L$ is the number of labelled samples, $X_i^L \in \mathbb{R}^{{n_g} ~ \text{x} ~ h}$ the input signal with $n_g$ sensors and $h$ time steps, $y_i^L \in \mathbb{R}$ the corresponding RUL label. The labelled set $D_L$ refers to strain measurement sequences of structures until failure. 
\end{enumerate}

Note that the samples of both domains $D_U$ and $D_L$ are multivariate time series sampled from related distributions.\\

\begin{figure}[!htp]
\centering
\includegraphics[scale=0.8]{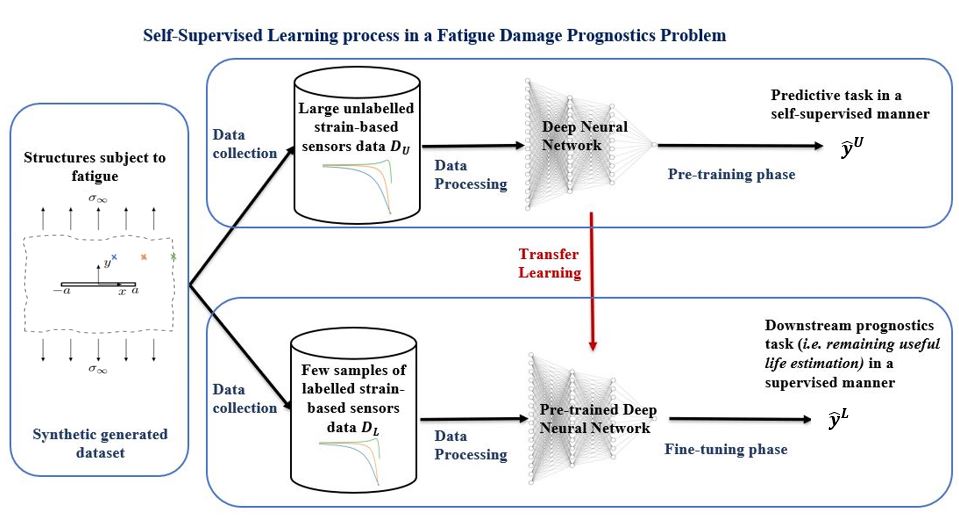}
\caption{Flow chart of the proposed Self-Supervised Learning framework.}
\label{fig:fc}
\end{figure}

Therefore in this paper, the pre-training phase of the proposed SSL approach consists of pre-training a DL model on unlabelled sensors dataset $D_U$ in a self supervised manner, called \textit{pre-text task}. 
The pre-trained model is then fine-tuned on a specific downstream Prognostics task, \textit{i.e.} RUL estimation, using only limited amounts of labelled data (\textit{i.e.} strain data of structures until failure, on which the RUL is known at each timestep). The pre-trained model is then fine-tuned on a specific downstream Prognostics task, \textit{i.e.} RUL estimation, using only limited amounts of labelled data $D_L$.
In this work, Deep Gated Recurrent Unit (GRU \cite{rana2016gated}) networks, or DGN, are used as the basic deep prediction model, because of their sequential properties and good regressive performance found in previous work \cite{ICPHM}) (see \ref{GRU} for more details about the GRU networks). Note that a DGN consists of a stack of GRU layers in this work.
Fig. \ref{fig:fc} summarizes the proposed SSL approach in this paper.

\subsubsection{Pre-training phase}

In order to vary the pre-text tasks and inspired by \cite{liu2021self, chen2020generative}, two types of models are used and compared in this work: 1)Autoencoders (AE) and 2)Autoregressive (AR) models.

\paragraph{Autoencoder architecture in pre-training phase}

\begin{figure}[!htp]
\centering
\includegraphics[scale=0.5]{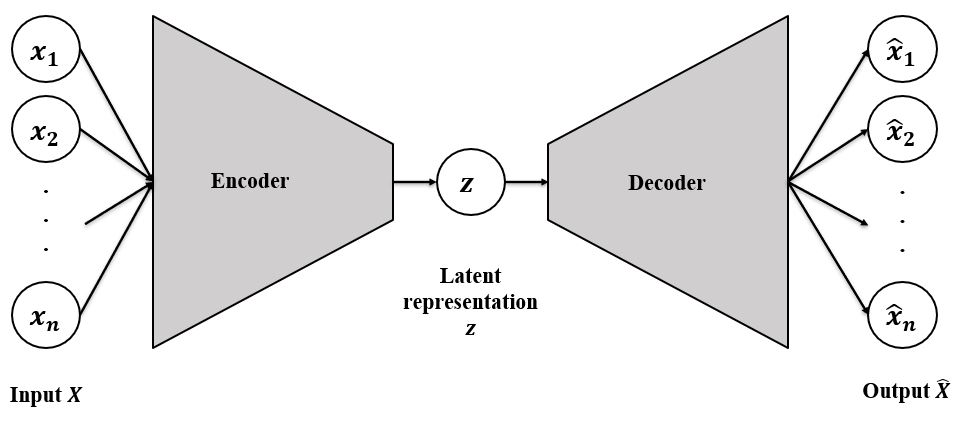}
\caption{The architecture of basic Autoencoders. Note that encoders and decoders can be composed of one or more hidden layers.}
\label{fig:AE_simp}
\end{figure}

An Autoencoder (AE) is an artificial neural network that is often used in learning the discriminating features of a dataset in an unsupervised manner \cite{rumelhart1986learning}. It is composed of two blocks: encoder and decoder (see Fig. \ref{fig:AE_simp} for a simplified architecture of the model). The encoder seeks to learn the underlying features of the input data $X_t$ at time step $t$. These learned features $z_t$ are generally of reduced dimension (number of neurons less than the number of input features). The goal of the decoder is thus to recreate the original data from these underlying learned features.
In recent years, Autoencoders have been successful in prognostics applications in terms of feature extraction \cite{ren2018bearing,ma2018predicting,sun2018deep}, which motivated the use of its architecture as a reference model for abstract representation learning in this work.\\

\begin{figure}[!htp]
\centering
\includegraphics[scale=0.7]{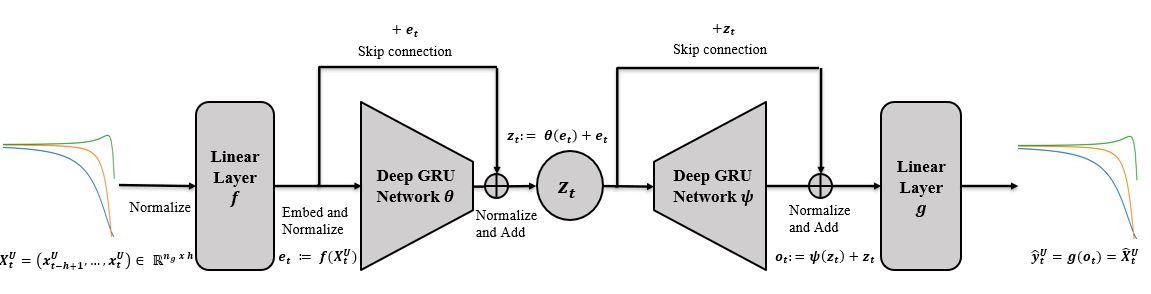}
\caption{Flow chart of the pre-training phase of the Autoencoder model (AE) in the proposed Self-Supervised Learning framework.}
\label{fig:SSLAE1}
\end{figure}

In pre-training, the output of the Autoencoder (AE) is an estimation of the unlabelled input signal $X^U_t = (x^U_{t-h+1}, ..., x^U_{t})$ such that $y^U_t = X_t$. A schematic view of the investigated AE model in the proposed SSL framework is given in Fig. \ref{fig:SSLAE1}.

The architecture of the AE model is organized as follows:

\begin{enumerate}
    \item The input data $X^U_t$ is first embedded through a linear layer\footnote{The input linear layer is used as an alternative to the embedding layers used in Natural Language Processing \cite{hrinchuk2019tensorized} since the input data is continuous is this work, converting each time step data into a fixed length vector of defined size.} $f$ in order to expand the dimension of the data and learn abstract features;
    
    \item The output of the following layer $e_t$ corresponds to a normalized\footnote{The layer normalization \cite{ba2016layer} are used for regularized training and faster convergence} transformation of the embedded input $f(X^U_t)$;
    \item The resulting embedded and normalized transformation of the data $e_t$ is then fed to an encoder $\theta$ and decoder $\psi$. Note that both encoder and decoder are Deep GRU networks (DGN), \textit{i.e.} stack of GRU layers;
    \item $z_t$ is considered as the learned representation by the model and will be used for feature extraction in the following;
    \item In this architecture, two skip connections\footnote{Skip connections in DL architectures, also called \textit{residual connections} or \textit{shortcut connections}, consist in skipping some layers in the neural network and feeding the output of one layer as the input to the next layers \cite{adaloglou2020intuitive}, used to solve the \textit{degradation problem} (e.g. ResNet \cite{he2016deep}). In this paper, skip connections are proposed to establish a direct connection through deep GRU networks in order to avoid information loss and learn robust sequential representation, which has already proven to be effective for deep recurrent neural networks \cite{yue2018residual}} are used through deep GRU networks such that $z_t = \theta(e_t) + e_t$ and $o_t = \psi(z_t) + z_t$;
    
    \item The output linear layer $g$ is then used to generate an estimation $\hat{y}^U_t$ of the unlabelled input signal $X^U_t$, \textit{i.e.} an estimation of the input signal such that $\hat{y}^U_t = \hat{X}^U_{t}$.
\end{enumerate}

\paragraph{Autoregressive architecture in pre-training phase}

\begin{figure}[!htp]
\centering
\includegraphics[scale=0.7]{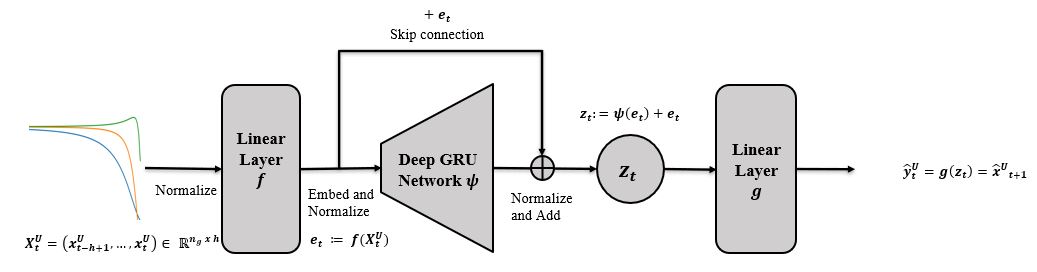}
\caption{Flow chart of the pre-training phase of the Autoregressive model (AR) in the proposed Self-Supervised Learning framework.}
\label{fig:SSLAR1}
\end{figure}

An Autoregressive (AR) model $g_h$ can be defined as a sequential model governed by an Autoregressive process of order $h$ that models the future outcome of a sequence at time $t+1$, using its previous $h$ realizations. Autoregressive modeling captures the temporal dependencies between sequential input data, which makes it useful in learning better features. 
Inspired by the autoregressive DL models used in \cite{brown2020language, chen2020generative} for pre-training, the proposed AR model in this paper consists of a Deep GRU network in which the input is a sequence of $h$ time steps at $t$ such that $X^U_t = (x^U_{t-h+1}, ..., x^U_{t})$, and the output is an estimation of the data of the next timestep such that $y^U_t = x^U_{t+1}$, according to the following formula:

\begin{equation}
\begin{split}
\begin{aligned}
\hat{y}^U_t &= \hat{x}^U_{t+1} = \mathcal{F}_h(x^U_{t}, x^U_{t-1}, \dots , x^U_{t-h+1})
\end{aligned}
\end{split}
\end{equation}
where $\mathcal{F}_h$ denotes the AR model governed by an autoregressive process of order $h$. A schematic view of the investigated AR models in the proposed SSL framework is given in Fig. \ref{fig:SSLAR1}. \\

The architecture of the AR model is organized as follows:
\begin{enumerate}
    \item The embedding linear layer $f$ is used in order to expand the dimension of the input data and learn abstract features;
    \item The output of the following layer $e_t$ corresponds to a normalized transformation of the embedded input $f(X^U_t)$;
    \item The resulting embedded and normalized transformation of the data $e_t$ is then fed to a Deep GRU Network $\psi$, composed of a stack of GRU layers;
    \item $z_t$ is considered as the learned representation by the model and will be used for feature extraction in the following;
    \item In this architecture, a skip connection is used such that $z_t = \psi(e_t) + e_t$;
    \item The output linear layer $g$ is then used to generate an estimation $\hat{y}^U_t$ of the input signal $X^U_t$, \textit{i.e.} the data of the next timestep such that $\hat{y}^U_t = \hat{x}^U_{t+1}$.
\end{enumerate}

\subsubsection{Fine-tuning phase}

\begin{figure}[!htp]
\centering
\includegraphics[scale=0.7]{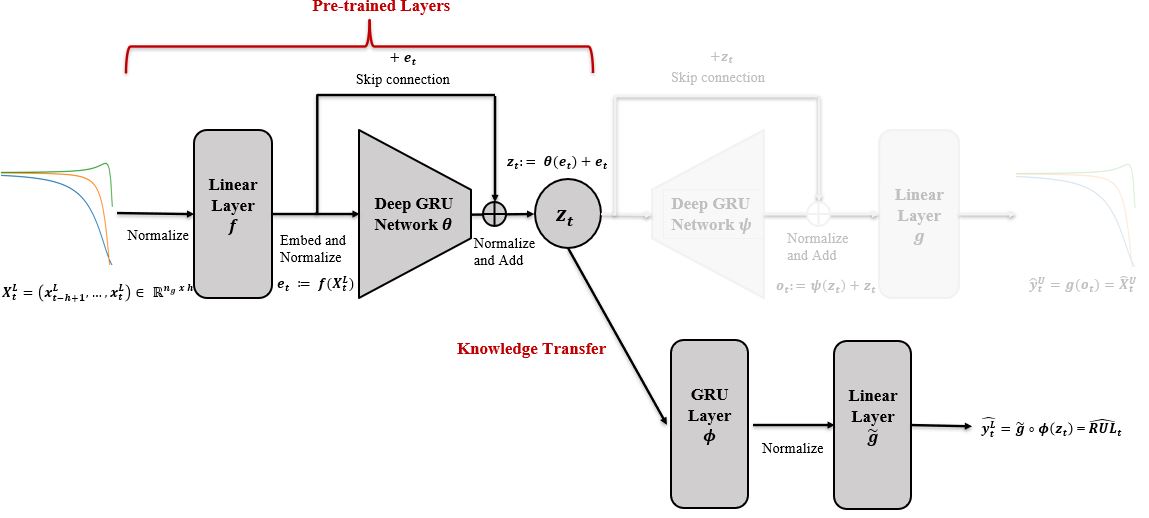}
\caption{Flow chart of the fine-tuning phase of the Autoencoder model (AE) in the proposed Self-Supervised Learning framework.}
\label{fig:SSLAE2}
\end{figure}

\begin{figure}[!htp]
\centering
\includegraphics[scale=0.7]{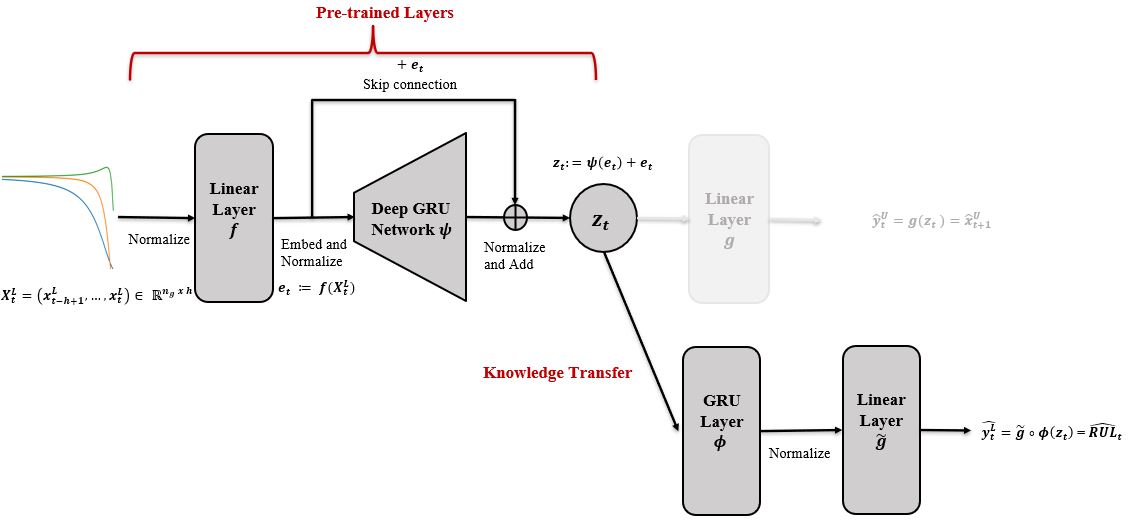}
\caption{Flow chart of the fine-tuning phase of the Autoregressive model (AR) in the proposed Self-Supervised Learning framework.}
\label{fig:SSLAR2}
\end{figure}

In the fine-tuning phase, an RUL estimation problem is considered, hence the output of the predictive models is a point-wise estimation of the RUL such that $y^L_t = RUL_t$. The embedding $z_t$ of the input data is extracted (see Fig. \ref{fig:SSLAE2} and Fig. \ref{fig:SSLAR2}), the weights of the hidden pre-trained layers are frozen, then for fine-tuning a simple GRU layer $\phi$ followed by an output linear layer $\Tilde{g}$ are used such that:

\begin{equation}
\begin{split}
\begin{aligned}
\hat{y}^L_t &= \Tilde{g} \circ \phi(z_t) \\
&= \hat{RUL}_t \\
\end{aligned}
\end{split}
\end{equation}
where the function $\phi$ refers to the fine-tuning GRU layer and the function $\Tilde{g}$ to the output linear layer. Note that, in the fine-tuning phase it is common to use only a linear layer for training, but the authors found that adding a GRU layer significantly improves the performance of the approach on this RUL estimation problem.\\

\newpage

Finally, in order to investigate the added value of the SSL approach in prognostics, the pre-trained models are compared with their non-pre-trained counterpart architecture, illustrated in Fig. \ref{fig:SSLAE2} and Fig. \ref{fig:SSLAR2}. Note that the architectures of the pre-trained and non-pre-trained models are the same, the difference residing in the absence of pre-training on unlabelled data and the corresponding \textit{knowledge transfer}. Also, only the non-pre-trained weights for the pre-trained models are trained (\textit{i.e.} trainable weights of the fine-tuning model), while all the trainable parameters of the non-pre-trained models are trained. Note that the pre-trained model with an autoregressive pre-text task followed by a GRU model for fine-tuning will be referred to as the ``autoregressive model" in the following for simplicity.\\

\section{Experiments and Results}
\label{sec:exp}

\subsection{Preparation of Data}

In this experiment, an Aluminum alloy 7075-T6 plate was considered, which is typical of aeronautic structures. Considering that the evolution of the changes from one cycle to another are small (see Fig. \ref{fig:fig33}), it was decided to collect the data every $\Delta{k} = 500$ loading-unloading cycles, as in \cite{ICPHM}. 

In the current paper, a training set and a testing set are generated. The training set is composed of:
\begin{enumerate}
    \item $N_U^{Train}$ unlabelled structures for the pre-training phase,
    \item $N_L^{Train}$ labelled structures for the fine-tuning phase.
\end{enumerate} 
Note that the number of structures $N_U^{Train}$ and $N_{L}^{Train}$ are varied; this will be described in the following subsections.
The testing set is composed of $N_L^{Test}$ labelled structures. It is used to evaluate the RUL estimation performance of the trained models (in fine-tuning) as a data set that was not used during training. The parameters used to generate the dataset according to the framework described in \cite{ICPHM} are summarized in Table \ref{tab:material_param}. \\

\begin{table}[!htp]

  \hspace{-2.5cm}
  \footnotesize
  \begin{tabular}{l c c c c}
  Parameter & Denotation & Type & Value & Unit\\
  
    \hline
    \textbf{Elastic parameters} & & & &  \\
    Young's modulus & $E$ & Deterministic & 71.7 & $GPa$ \\
    Poisson's ratio & $\nu$ & Deterministic & 0.33 & -\\
    \hline
    \textbf{Strain field parameters} & & & &  \\
    Maximum stress intensity & $\sigma_{max}$ & Uniform distribution & $\mathcal{U}(75,85).10^6$  & $Pa$\\
    Fracture toughness & $K_{I}$ & Deterministic & $19,7.10^6$  & $Pa\sqrt{m}$\\

    \hline
    \textbf{Strain gauges} & & & &  \\
    Number of gauges placed & $n_g$ & Deterministic & 3 & - \\
    Position of the gauges placed & $(x_i,y_i)_{i=1,..,n_g}$ & Deterministic & $(3,14), (14,14), (25,14)$  & $mm$\\
    Angle of the gauges placed & $\theta$ & Deterministic & $45$  & $deg$ \\
    
    \hline
    \textbf{Initialization parameters} & & & &  \\
    Initial crack size & $a_0$ & Gaussian distribution & $\mathcal{N}(\mu_{a_0}, \sigma_{a_0})$  & $m$\\
    Mean of $a_0$ & $\mu_{a_0}$ & Deterministic & $5.10^{-4}$  & $m$\\
    Standard deviation of $a_0$ & $\sigma_{a_0}$ & Deterministic & $2,5.10^{-4}$  & $m$\\
    
    Paris-Erdogan's law parameters & ($m, \log{C}$) & Multivariate Gaussian distribution & $\mathcal{N}(\mu_m, \sigma_m, \mu_{\log{C}}, \sigma_{\log{C}}, \rho)$ & -\\
    
    Mean of $m$ & $\mu_{m}$ & Deterministic & $3,4$  & -
    \\
    Standard deviation of $m$ & $\sigma_{m}$ & Deterministic & $0,25$  & -\\
    
    Mean of $C$ & $\mu_{C}$ & Deterministic & $1.10^{-10}$  & -\\
    Standard deviation of $C$ & $\sigma_{C}$ & Deterministic & $5.10^{-11} $  & -\\
    
    Correlation coefficient of $m$ and $\log{C}$ & $\rho$ & Deterministic & $-0.996 $  & -\\
    \hline
    \textbf{Generated data set} & & & &  \\
    Number of unlabelled structures for training & $N_{U}^{Train}$ & Deterministic & $(100, 1000, 5000, 10000)$  & -\\
    Number of labelled structures for training & $N_{L}^{Train}$ & Deterministic & $(5, 10, 20, 50, 100)$  & -\\
    Number of labelled structures for testing & $N_L^{Test}$ & Deterministic & $100$  & -\\
    Data collection interval & $\Delta k$ & Deterministic & $500$  & -\\
    
    \hline
  \end{tabular}
  \caption{Parameters for numerical study. The full model description is available in \cite{ICPHM}.}
  \label{tab:material_param}
\end{table}
\newpage

\subsection{Experimental settings in pre-training phase}

\begin{figure}[!htp]
\centering
\includegraphics[width=120mm]{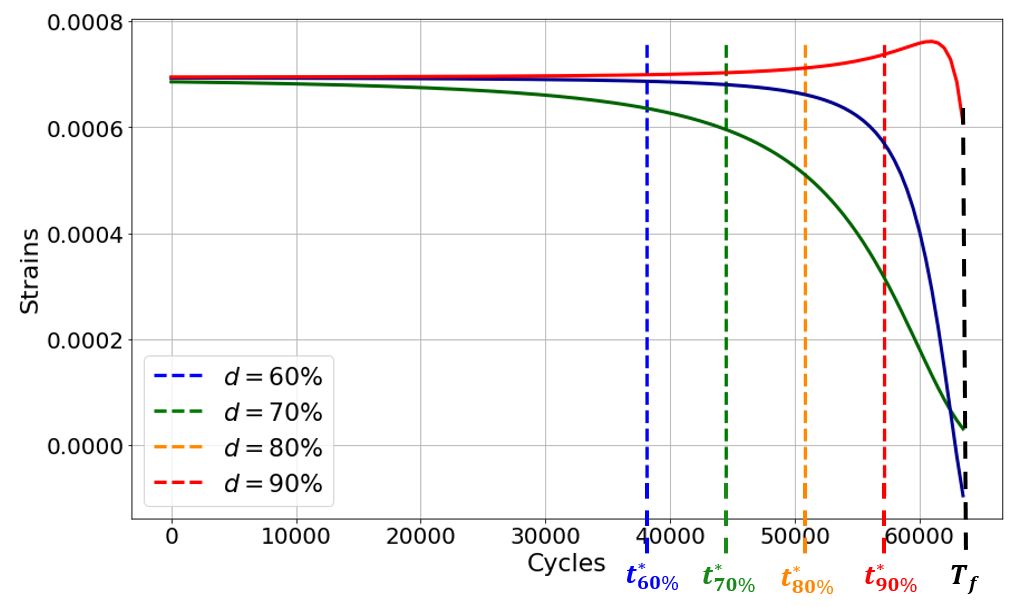}
\caption{\footnotesize Four degradation scenarios depicted on the sequence of a structure. For each scenario, the available strain data correspond to the measurements from time $0$ to time $t^*_d:= d \text{~x~} T_f$, where $d$ is the ratio of the total lifetime of a sequence, and $T_f$ is the time of failure.}
\label{fig:d}
\end{figure}

As the structures subjected to fatigue can be replaced before reaching failure at any time, the proposed approach has been investigated on four degradation scenarios: for pre-training, available sequences of unlabelled data are incomplete at \textit{d} = 60\%, 70\%, 80\%, and 90\% of their total lifetime, where $d$ is the ratio of the total lifetime of a sequence. To illustrate the size of the strain data sequences available, these four degradation scenarios are illustrated in Fig. \ref{fig:d}. \\

% \newpage
Moreover, the number of pre-training structures, denoted $N_U^{Train}$, for which strain sequences were available was varied in order to investigate the effect of the amount of unlabelled data. The investigated models (autoencoder and autoregressive model) were therefore pre-trained on $N_U^{Train}$ $= 100, 1000, 5000,$ and $10000$ unlabelled structures. As mentioned before, strain data are collected every 500 cycles, and a sliding window approach of $h = $ 30 is used (see Section \ref{sec:approach}). Thus, as an illustration, Table \ref{tab:AE_samples} summarises the number of pre-training samples $n_U$ for the autoencoder model in each degradation scenario.\\

	% \diagbox[width=15em]{Number of \\pre-training structures \\$N_{U}^{Train}$}{Ratio of the\\ total lifetime  $d$}  & $d$ = $60\%$ & $d$ = $70\%$& $d = 80\%$ & $d = 90\%$\\

\linespread{1}
\begin{table}[!htp]
    \footnotesize
    
	\centering
	\setlength{\tabcolsep}{5pt}
	\begin{tabular}{|c|l|l|l|l|c|l|}
	    \hline
		\diagbox[width=10em]{Number \\of pre-training  \\structures $N_{U}^{Train}$\vspace{0.1cm}}{Ratio of the\\ total lifetime  $d$}  & $d$ = $60\%$ & $d$ = $70\%$& $d = 80\%$ & $d = 90\%$\\
		\hline
        $N_{U}^{Train} = 100$ & $n_U^{Train}$ =  11880&  $n_U^{Train}$ = 14346&  $n_U^{Train}$ = 16819&  $n_U^{Train}$ = 19283\\
        $N_{U}^{Train} = 1000$ &  $n_U^{Train}$ = 114537&  $n_U^{Train}$ = 138451&  $n_U^{Train}$ = 162511&  $n_U^{Train}$ = 186443\\
        $N_{U}^{Train} = 5000$ &  $n_U^{Train}$ = 571515&  $n_U^{Train}$ = 690932&  $n_U^{Train}$ = 811015&  $n_U^{Train}$ = 930545\\
        $N_{U}^{Train} = 10000$ &  $n_U^{Train}$ = 1137959&  $n_U^{Train}$ = 1375948&  $n_U^{Train}$ = 1615261& $n_U^{Train}$ = 1853470\\
		\hline
		
	\end{tabular}
	\caption{\footnotesize Number of unlabelled pre-training samples $n_U^{Train}$ used in this work for the autoencoder. Note that in this work, strain data are collected every $\Delta_k = 500$ cycles and a sliding window approach of $h = 30$ is used.}
	\label{tab:AE_samples}
\end{table}

In each training procedure, 95\% of the dataset was used for training (in terms of the number of structures), while 5\% of it was used for validation. The validation set is used for monitoring and adjusting the training phase, using the mean absolute percentage error (\textit{MAPE}) metric. During training, the aim is to minimise the mean squared error (\textit{MSE}) loss function $L_{MSE}$ such that:

\begin{eqnarray}
	\label{eq:accuracy_metric}
    L_{MSE}  &=& \frac{1}{n_U}\sum_{i=1}^{n_U} (y^U_i - \hat{y}^U_i)^2 \\
	\mathrm{\textit{MAPE}}  &=& \frac{1}{n_U}\sum_{i=1}^{n_U} |\frac{y^U_i - \hat{y}^U_i}{y^U_i}| * 100 
\end{eqnarray}
\noindent where $ n_U $ is the number of unlabelled samples with $ \hat{y}^U_{.} $ being the prediction and $ y^U_{.} $ the target value. Note that $y^U_i = (x^U_{i-h+1}, ..., x^U_{i})$ for the AE model and $y^U_i = x^U_{i+1}$ for the AR model. The Adam optimizer \cite{kingma2014adam} was used with default parameters and the learning rate was decreased incrementally. The learning rates of $10^{-2}, 10^{-3}, 10^{-4}$ were sequentially used for a predefined number of epochs, saving the model weights each time the validation loss decreases; the weights of the best model were loaded each time the learning rate was lowered. At the end of the procedure, the model was trained on the whole dataset (training and validation sets) with a lower learning rate of $10^{-5}$ until convergence.  Calculations were performed using PyTorch's core library in Python on NVIDIA V100 GPUs, hence the batch size was chosen to be as large as possible in order to speed up calculations, here $2^{12}$ = 4096 depending on the available memory of the used GPUs, and not too large in order to avoid numerical instability. The model hyperparameters were optimized using a Grid Search algorithm, listed in Table \ref{tab:hp_pt}:

\begin{itemize}
    \item Autoencoder model illustrated in Fig. \ref{fig:SSLAE1}: the embedding linear layer $f$ is composed of 64 neurons, the Deep GRU Networks $\theta$ and $\psi$ were each composed of 2 layers of GRU, 64 neurons, and a dropout of 0.1, that is to say nearly 100.000 parameters.
    \item Autoregressive model illustrated in Fig. \ref{fig:SSLAR1}: the embedding linear layer $f$ is composed of 64 neurons, the Deep GRU Network $\psi$ was composed of 4 GRU layers, 64 neurons, and a dropout of 0.1, that is to say nearly 100.000 parameters.
\end{itemize}
% \newpage

\linespread{1}
\begin{table}[!htp]
    \footnotesize

	\centering
	\setlength{\tabcolsep}{5pt}
	\begin{tabular}{l| c c c}
	    \hline
		\textbf{Hyperparameters} & \textbf{Search Space} & \textbf{Autoencoder model} & \textbf{Autoregressive model}\\
		\hline
		\hline
        Linear layer $f$ - neurons & \{32, 64\} & 64 & 64\\
        Deep GRU network $\theta$ - neurons & \{32, 64, 128, 256\} & 64& - \\
        Deep GRU network $\theta$ - layers & \{1, 2, 4, 8\} & 2& - \\
        Deep GRU network $\theta$ - dropout & \{0, 0.1, 0.2, 0.3\} & 0.1& - \\
        Deep GRU network $\psi$ - neurons & \{32, 64, 128, 256\} & 64&64\\
        Deep GRU network $\psi$ - layers & \{1, 2, 4, 8\}& 2&4\\
        Deep GRU network $\psi$ - dropout & \{0, 0.1, 0.2, 0.3\} & 0.1&0.1\\

		\hline
		
	\end{tabular}
	\caption{\footnotesize Hyperparameters of the pre-training phase.}
	\label{tab:hp_pt}
\end{table} 

Note that the authors found that, given the training data, the search space considered was sufficient to obtain good results, whereas deep neural networks with a larger number of layers/neurons performed poorer with longer training (probably due to more difficult convergence).

\subsection{Experimental settings in Fine-tuning phase}

For fine-tuning, as illustrated in Fig. \ref{fig:SSLAE2} and Fig. \ref{fig:SSLAR2}, the embedding $z_t$ of the input data was extracted, the weights of the hidden layers were frozen, and a GRU model was used for the downstream task. The models are then trained on $N_{L}$ available labelled structures, using a sliding window approach similar to that used in the previous pre-training phase. The fine-tuning model was composed of a single GRU layer, 32 neurons, 0.1 in dropout to regularize, and followed by an output linear layer (the hyperparameters were optimized using a Grid Search algorithm on the autoencoder pre-trained model, listed in Table \ref{tab:hp_ft}). \\

\linespread{1}
\begin{table}[!htp]
    \footnotesize

	\centering
	\setlength{\tabcolsep}{5pt}
	\begin{tabular}{l| c c}
	    \hline
		\textbf{Hyperparameters} & \textbf{Search Space} & \textbf{Fine-tuning model}\\
		\hline
		\hline
        Deep GRU network $\phi$ - neurons & \{32, 64\} & 32 \\
        Deep GRU network $\phi$ - layers & \{1, 2\} & 1 \\
        Deep GRU network $\phi$ - dropout & \{0, 0.1, 0.2, 0.3\} & 0.1 \\
        Batch size & \{32, 64\} & 32 \\

		\hline
		
	\end{tabular}
	\caption{\footnotesize Hyperparameters of the fine-tuning phase.}
	\label{tab:hp_ft}
\end{table} 

The pre-trained models were then compared with their non-pre-trained ``counterpart" (\textit{i.e.} same architecture but all model weights were reset) on few shots learning. 
The number of available labelled training structures $N_{L}^{Train}$ were varied, such that: $N_{L}^{Train} = 5, 10, 20, 50 $ and $ 100$ labelled structures (\textit{i.e.} strain data of structures reaching failure at time $T_f$, thus for which the RUL is available for each timestep between times 0 and $T_f$). Calculations were performed using PyTorch's core library in Python on a machine with 62GB of RAM and an NVIDIA GeForce GTX 1080 Ti 11 GB GPU.\\

After training during the fine-tuning phase, the models are evaluated on the testing set composed of $N_L^{Test} = 100$ different labelled structures. For each structure, a unique RUL estimation is performed at a time $t^*_n$. For each structure $n \in \{1,\dots, N_L^{Test}\}$, the parameter $t^*_n$ is randomly drawn such that $t^*_n \sim T_{f}^n \times \mathcal{U}([0,33 ; 0,9])$, where $T_{f}^n$ is the time of failure for the $n$-th structure. This means that the test prediction for the RUL is done at a time $t^*_n$ which is drawn uniformly between 33\% and 90\% of the sequence's length. Hence, the input data for the model is $X^L_{t^*_n} = (x^L_{t^*_n-h+1}, \dots, x^L_{t^*_n}) \in \mathbb{R}^{n_g ~ \text{x} ~ h}$ and the output of the model is $\hat{y}^L_{t^*_n} = \hat{RUL}_{t^*_n} \in \mathbb{R}$. As the RUL estimation problem is considered as a regression problem in this paper, the aim is to minimize a mean squared error loss $L_{MSE}$ during training, and the mean absolute percentage error (\textit{MAPE}) metric is used to evaluate the performance of the investigated models such that:

\begin{eqnarray}
    L_{MSE}  &=& \frac{1}{n_L}\sum_{i=1}^{n_L} (y^L_i - \hat{y}^L_i)^2 \\
	\mathrm{\textit{MAPE}}  &=& \frac{1}{n_L}\sum_{i=1}^{n_L} |\frac{y^L_i - \hat{y}^L_i}{y^L_i}| * 100 
\end{eqnarray}

\noindent where $ n_L $ is the number of labelled samples with $ \hat{y}^L_{.} $ being the RUL prediction and $ y^L_{.} $ the target RUL value.\\

\begin{figure}[!htp]
\centering
\includegraphics[width=160mm]{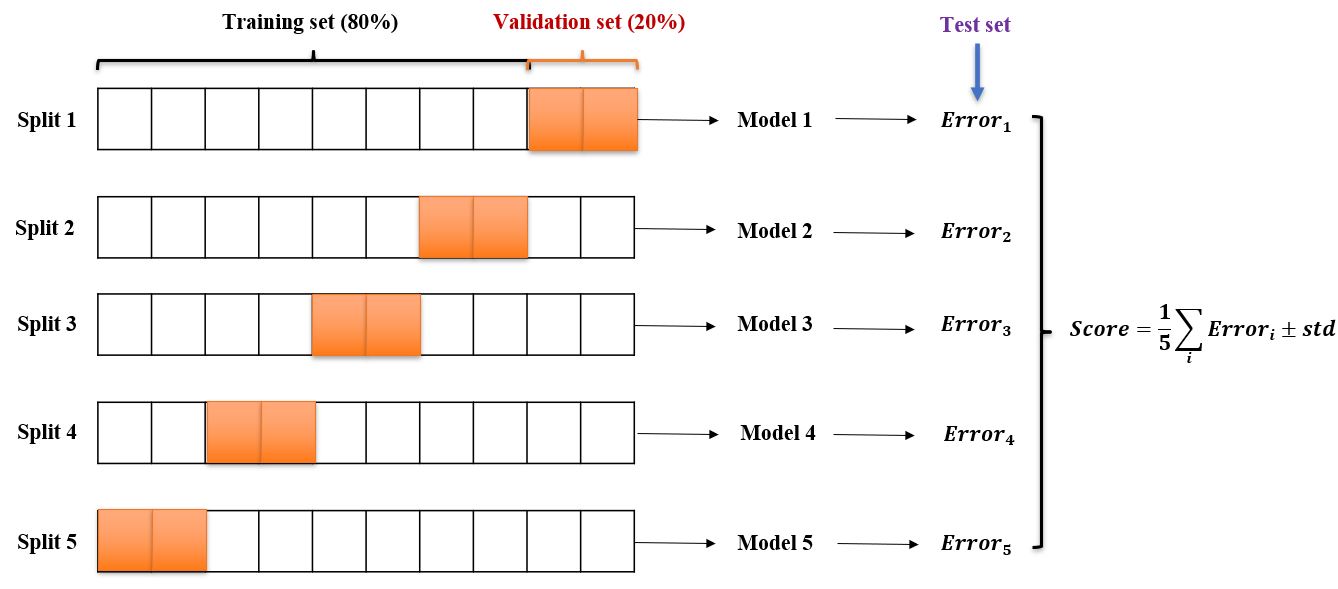}
% \captionsetup{labelformat=empty}
% \vspace{-0.32cm}
\caption{Flow chart of the 5-fold cross validation used in this paper.}
\label{fig:kfold}
\end{figure}

As a limited amount of labelled data leads to epistemic uncertainty, it is difficult to make a reliable comparison. Hence, a 5-fold cross validation was used by varying the split between the training and validation set, as illustrated in Fig. \ref{fig:kfold}, which gives an average \textit{MAPE} error and its standard deviation to quantify the uncertainty when evaluated on the test set.\\

\subsection{Results}

\subsubsection{Pre-training analysis and comparison of pre-text tasks}

First, the performance of the pre-trained Autoencoder (AE) and non-pre-trained counterpart were compared on the considered RUL estimation problem, by varying the number of unlabelled samples in pre-training. The results are presented in Table \ref{tab:mapeAE}. \\

\linespread{1}
\begin{table}[!htp]
    \footnotesize
   
	\centering

	\setlength{\tabcolsep}{5pt}
	\begin{tabular}{l|llllllll}
	    \hline
		\hline
		  & &  $MAPE$ (\%) & Mean  $\pm$ St. & Dev.\\
		\hline
		\hline
		Labelled structures $N_{L}^{Train}$ & $5$ & $ 10$& 20 & 50 &  100\\
		\hline

		\textbf{Pre-trained model ($d$ = 60\%)} & & & \\
		Autoencoder $N_U^{Train}$ = 100 & 29.36 $\pm$ 3.81 & 21.16 $\pm$ 2.86 & 11.84 $\pm$ 4.12 &$1.95 \pm 0.25$ & $1.87 \pm 0.36$\\
		Autoencoder $N_U^{Train}$ = 1.000 & 25.95 $\pm$ 0.74 & 14.78 $\pm$ 3.13 & 5.59 $\pm$ 1.47  & $1.42 \pm 0.17$ & $1.31 \pm 0.06$\\ 
		Autoencoder $N_U^{Train}$ = 5.000 & 28.55 $\pm$ 0.99 & \textbf{12.01} $\pm$ \textbf{2.31} & 4.80 $\pm$ 0.92 & \textbf{1.27 $\pm$ 0.11} & \textbf{1.08 $\pm$ 0.03}\\ 
		Autoencoder $N_U^{Train}$ = 10.000 & \textbf{24.60} $\pm$ \textbf{4.30}  & 12.70 $\pm$ 1.76 & \textbf{3.48} $\pm$ \textbf{0.93} & 1.41 $\pm$ 0.04 & 1.13 $\pm$ 0.12\\

		\hline
		
		\textbf{Pre-trained model ($d$ = 70\%)} & & & \\
		Autoencoder $N_U^{Train}$ = 100 & 28.02 $\pm$ 1.96 & 20.99 $\pm$ 3.73 & 11.70 $\pm$ 3.78 & $1.61 \pm 0.43$ & $1.43 \pm 0.25$\\
		Autoencoder $N_U^{Train}$ = 1.000 & 30.76 $\pm$ 1.83 & $18.20 \pm 3.76$ & $6.42 \pm 1.69$ & $1.65 \pm 0.36$ & $1.29 \pm 0.16$\\ 
		Autoencoder $N_U^{Train}$ = 5.000 & 26.10 $\pm$ 0.88 & $18.73 \pm 2.72$ & $6.67 \pm 1.30$ & $1.58 \pm 0.23$ & $1.25 \pm 0.08$\\ 
		Autoencoder $N_U^{Train}$ = 10.000 & \textbf{25.38} $\pm$ \textbf{5.99}   & \textbf{11.85} $\pm$ \textbf{2.64} & $\textbf{3.77} \pm \textbf{0.61}$ & $\textbf{1.29} \pm \textbf{0.06}$ & $\textbf{1.08} \pm \textbf{0.12}$ \\
		\hline
		
		\textbf{Pre-trained model ($d$ = 80\%)} & & & \\
		Autoencoder $N_U^{Train}$ = 100 & 34.96 $\pm$ 9.59 & 17.72 $\pm$ 4.43 & 8.92 $\pm$ 3.33 & $1.33 \pm 0.13$ & $1.41 \pm 0.20$\\
		Autoencoder $N_U^{Train}$ = 1.000 & 26.46 $\pm$ 2.51 & 15.78 $\pm$ 3.77 & 4.74 $\pm$ 0.79  & 1.31 $\pm$ 0.13 & $1.10 \pm 0.05$\\ 
		Autoencoder $N_U^{Train}$ = 5.000 & 30.19 $\pm$ 6.56 & 17.18 $\pm$ 3.03 & 6.19 $\pm$ 2.07 & 1.60 $\pm$ 0.22 & $1.17 \pm 0.18$\\ 
		Autoencoder $N_U^{Train}$ = 10.000 & \textbf{24.77} $\pm$ \textbf{4.80}   & \textbf{12.06} $\pm$ \textbf{3.82} & \textbf{4.27} $\pm$ \textbf{0.72} & \textbf{1.18} $\pm$ \textbf{0.13} & \textbf{1.01} $\pm$ \textbf{0.06} \\
		\hline
		
		\textbf{Pre-trained model ($d$ = 90\%)} & & & \\
		Autoencoder $N_U^{Train}$ = 100 & 28.27 $\pm$ 2.47 & 21.56 $\pm$ 0.98 & 9.91 $\pm$ 3.29 & 1.57 $\pm$ 0.35  & 1.56 $\pm$ 0.16\\
		Autoencoder $N_U^{Train}$ = 1.000 & 30.27 $\pm$ 1.15 & 18.53 $\pm$ 5.68 & 6.51 $\pm$ 1.63 & 1.74 $\pm$ 0.43 & 1.13 $\pm$ 0.15\\ 
		Autoencoder $N_U^{Train}$ = 5.000 & 25.99 $\pm$ 1.73 & 17.06 $\pm$ 4.09 & 5.24 $\pm$ 1.60 & 1.43 $\pm$ 0.31 & 1.06 $\pm$ 0.07\\ 
		Autoencoder $N_U^{Train}$ = 10.000 & \textbf{22.97} $\pm$ \textbf{5.69}   & \textbf{11.04} $\pm$ \textbf{3.61}  & $\textbf{3.39} \pm \textbf{0.67}$ & \textbf{1.22}$\pm$ \textbf{0.12} & \textbf{0.88} $\pm$ \textbf{0.09} \\
		\hline

		\textbf{Non-pre-trained model} & & &  & \\
		Autoencoder architecture  &  27.72 $\pm$ 0.65   & 23.07 $\pm$ 5.94 &  8.12 $\pm$ 1.87 & 1.40 $\pm$ 0.33 & \textbf{0.83 $\pm$ 0.18} \\
		
	\end{tabular}
	\caption{\footnotesize \textit{MAPE} mean values (in \%) plus or minus its standard deviation as a function of the number of labelled structures used and of the training scenario for the autoencoder case. Best performance is represented in bold for each case.}
	\label{tab:mapeAE}
\end{table} 
% \newpage

A first remark that can be drawn from the results in Table \ref{tab:mapeAE} is that pre-training the model is not always beneficial. For example, for an AE model pre-trained on 100 structures, the accuracy of the RUL estimation is not always better compared to the non-pre-trained model (AE). The term \textit{negative transfer} can be used when the transfer method decreases predictive performance \cite{torrey2010transfer}. Moreover, considerable variability in results can be observed when models are pre-trained on very few unlabelled samples (for example when fine-tuned on 5 labelled samples), which could be due to over-fitting during pre-training. However, it can also be observed that as the number of unlabelled samples increases, the pre-training becomes more efficient and allows to have better results than a non-pre-trained model when few labelled structures are available, especially for the model pre-trained on 10000 structures. For example, the AE pre-trained models with $N_U^{Train} = 10000$ unlabelled structures and fine-tuned on $N_{L}^{Train} = 10$ structures has an $MAPE$ of about 11-12\% while the non-pre-trained model with the same $N_{L}^{Train} = 10$ structures has an $MAPE$ of about 23\%. Overall, results in Table \ref{tab:mapeAE} show that for the Autoencoder model, the number of unlabelled samples in pre-training matters: the more unlabelled samples, the more efficient the self supervised learning is for each of the 4 scenarios. The autoregressive model shows similar performances, presented in Table \ref{tab:mapeAR}. \\

\linespread{1}
\begin{table}[!htp]
    \footnotesize
	\centering
	\setlength{\tabcolsep}{5pt}
	\begin{tabular}{l|llllllll}
	
        \hline
		\hline
		 & &  $MAPE$ (\%) & Mean  $\pm$ St. & Dev.\\
		\hline
		\hline
		Labelled structures $N_{L}^{Train}$ & $ 5$ & $ 10$& 20 & 50 &  100\\
		\hline
		\textbf{Pre-trained model (d = 60\%)} & & & \\
		Autoregressive $N_U^{Train}$ = 100 & 36.15 $\pm$ 13.48 & 20.18 $\pm$ 5.19 & 12.17 $\pm$ 3.31 & $2.29 \pm 0.22$ & $1.70 \pm 0.19$\\
		Autoregressive $N_U^{Train}$ = 1.000 & 28.14 $\pm$ 3.20 & 16.65 $\pm$ 1.21 & 8.80 $\pm$ 3.13 & $1.48 \pm 0.18$ & $1.21 \pm 0.13$ \\
		Autoregressive $N_U^{Train}$ = 5.000 & 26.53 $\pm$ 1.57  &  13.58 $\pm$ 2.14  &  7.25 $\pm$ 3.09 & 1.22 $\pm$ 0.02 & $1.00 \pm 0.01$ \\
		Autoregressive $N_U^{Train}$ = 10.000 &  \textbf{22.48} $\pm$ \textbf{6.06} & \textbf{7.34} $\pm$ \textbf{0.95}  & \textbf{2.63} $\pm$ \textbf{0.80} & $\textbf{1.14} \pm \textbf{0.04}$ &  \textbf{1.03} $\pm$ \textbf{0.06} \\
		
		\hline
		
		\textbf{Pre-trained model (d = 70\%)} & & &  &  \\
		Autoregressive $N_U^{Train}$ = 100 & 28.95 $\pm$ 1.74 & 16.98 $\pm$ 2.96 & 11.85 $\pm$ 2.73  & 2.01 $\pm$ 0.20  & 1.51 $\pm$ 0.34\\
		Autoregressive $N_U^{Train}$ = 1.000 & 26.76 $\pm$ 2.49 & 16.34 $\pm$ 1.38 & 8.54 $\pm$ 1.89 & 1.53 $\pm$ 0.20 & 1.18 $\pm$ 0.16 \\
		Autoregressive $N_U^{Train}$ = 5.000 & 25.18 $\pm$ 2.70  &  10.96 $\pm$ 2.46  &  6.79 $\pm$ 2.51 & 1.33 $\pm$ 0.13 &  1.27 $\pm$ 0.21\\
		Autoregressive $N_U^{Train}$ = 10.000 & \textbf{24.43} $\pm$ \textbf{4.08} & \textbf{8.46} $\pm$ \textbf{1.53} &  \textbf{2.42} $\pm$ \textbf{0.53}& \textbf{1.20} $\pm$ \textbf{0.06} &  \textbf{1.14} $\pm$ \textbf{0.17} \\
		
		\hline
		
		\textbf{Pre-trained model (d = 80\%)} & & & & \\
		
		Autoregressive $N_U^{Train}$ = 100 & 30.47 $\pm$ 3.47 & 20.04 $\pm$ 2.59 & 10.68 $\pm$ 4.25 & 2.34 $\pm$ 0.85 & 1.48 $\pm$ 0.07\\
		Autoregressive $N_U^{Train}$ = 1.000 & \textbf{26.10} $\pm$ \textbf{2.14} & 13.66 $\pm$ 3.22 & 6.01 $\pm$ 1.45 & 1.44 $\pm$ 0.05 & 1.17 $\pm$ 0.03 \\
		Autoregressive $N_U^{Train}$ = 5.000 & 26.46 $\pm$ 2.46 & 12.60 $\pm$ 1.33 & 6.91 $\pm$ 1.28 & 1.38 $\pm$ 0.11 & 1.09 $\pm$ 0.05  \\
		Autoregressive $N_U^{Train}$ = 10.000 & 26.50 $\pm$ 2.58  &  \textbf{8.09} $\pm$ \textbf{3.28}  &  \textbf{2.39} $\pm$ \textbf{0.26} & \textbf{1.39} $\pm$ \textbf{0.20} & \textbf{1.07} $\pm$ \textbf{0.11} \\
		
		\hline
		
		\textbf{Pre-trained model (d = 90\%)} & & & &  \\
		Autoregressive $N_U^{Train}$ = 100 & 35.76 $\pm$ 7.41 & 17.64 $\pm$ 2.58 & 8.66 $\pm$ 2.26 & 1.60 $\pm$ 0.17 & 1.33 $\pm$ 0.19\\
		Autoregressive $N_U^{Train}$ = 1.000 & 25.00 $\pm$ 3.89 & 17.31 $\pm$ 1.72 & 8.59 $\pm$ 2.79 & 1.45 $\pm$ 0.15 &  1.29 $\pm$ 0.20\\
		Autoregressive $N_U^{Train}$ = 5.000 & 23.01 $\pm$ 3.23 & 12.97 $\pm$ 2.91 & 3.15 $\pm$ 0.45 & 1.21 $\pm$ 0.08 &  1.05 $\pm$ 0.03\\
		
		Autoregressive $N_U^{Train}$ = 10.000 & \textbf{22.69} $\pm$ \textbf{2.36} & \textbf{8.83} $\pm$ \textbf{1.61} &  \textbf{3.39} $\pm$ \textbf{0.40}  & \textbf{1.25} $\pm$ \textbf{0.07} & \textbf{0.99} $\pm$ \textbf{0.06} \\
		
		\hline

		\textbf{Non-pre-trained model} & & &  & \\
		Autoregressive architecture &  28.09 $\pm$ 1.63   & 21.10 $\pm$ 1.92 &  7.52 $\pm$ 1.59 & 1.15 $\pm$ 0.09 & \textbf{0.79} $\pm$ \textbf{0.09} \\

	\end{tabular}
	\caption{\textit{MAPE} mean values (in \%) plus or minus its standard deviation as a function of the number of labelled structures used and of the training scenario for the autoregressive case. Best performance is represented in bold for each case.}
	\label{tab:mapeAR}
\end{table}

\begin{figure}[!htp]
\centering
\includegraphics[width=140mm]{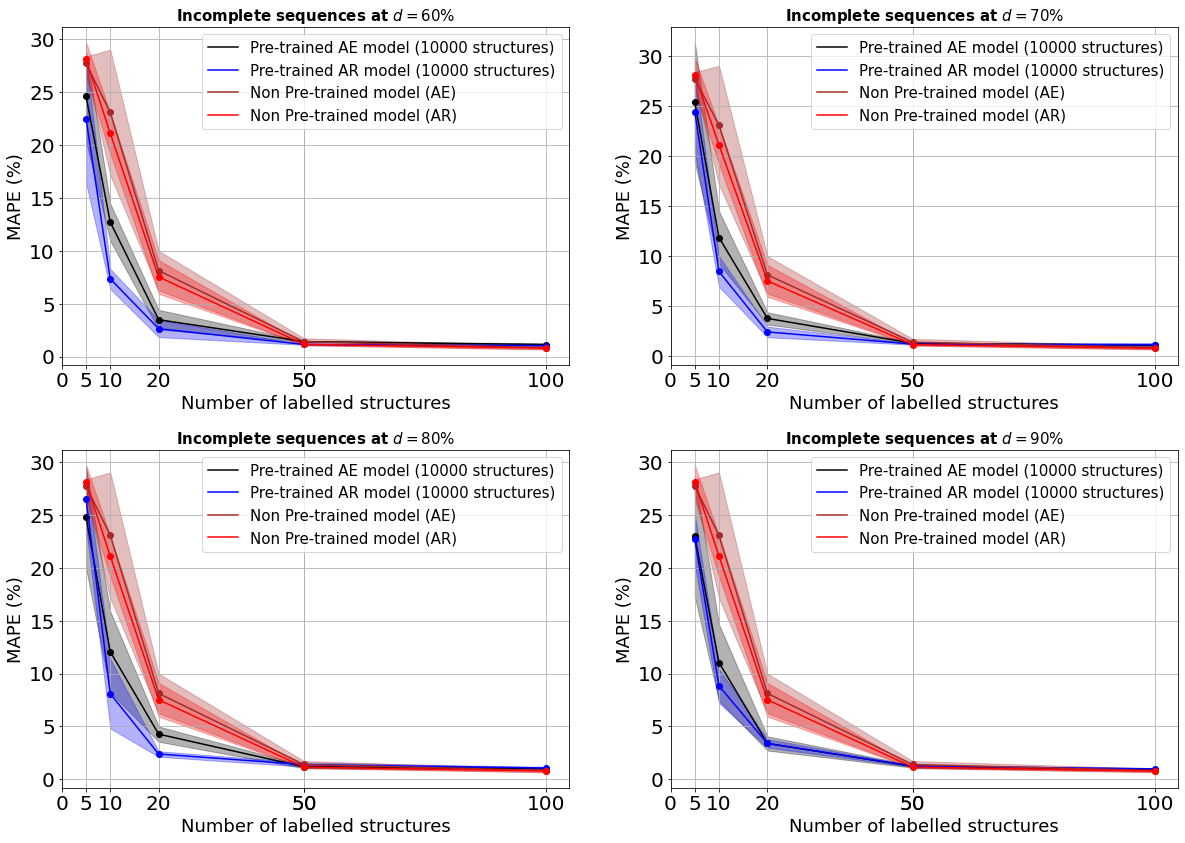}

\caption{Comparison of pre-trained and non-pre-trained models in RUL estimation for the testing set (100 samples). The \textit{MAPE} metric (\%) is used, and here the target value to estimate is the RUL. }
\label{fig:mapeALL}
\end{figure}

In order to illustrate the differences in performance between the autoencoder and the autoregressive pre-text tasks, the $MAPE$ of these models for $N_U^{Train} = 10000$ structures as well as the $MAPE$ of their non-pre-trained counterparts are provided in Fig. \ref{fig:mapeALL}. For very few labelled structures ($N_{L}^{Train} = 5$), results illustrated in Fig. \ref{fig:mapeALL} do not allow to clearly distinguish between the two models due to the limited number of labelled samples, leading all models to work relatively poorly.\\

Nevertheless, results show that both pre-trained models clearly outperform their non-pre-trained counterpart in Few-Shots learning (more than 5 but less than 50 structures). The AR pre-trained model significantly outperforms the AE pre-trained one when fine-tuned on 10 or 20 structures, and has almost three times less estimation error than the best non-pre-trained model. These results make sense since the autoregressive task and the RUL estimation task have in common the task of predicting future outcome, and may need to capture the temporal dependencies of the input signal.
However, it can also be seen that as the number of labelled samples increases, the difference between the pre-trained and non-pre-trained models is reduced (\textit{e.g. }trained on more than 50 structures). \\

\begin{figure}[!htp]
		\centering
        \vspace{-0.2cm}
		\includegraphics[width=0.7\textwidth]{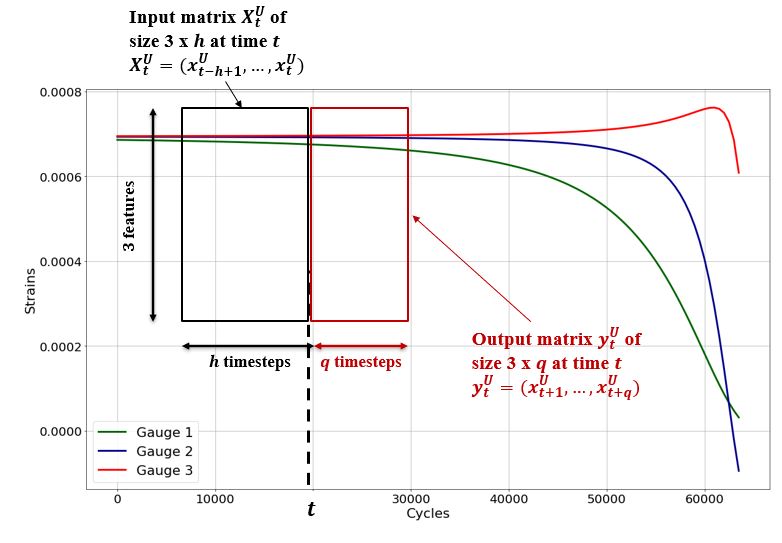}
		\caption{Illustration of the extended Autoregressive model (MSPA) pre-text task.}
	   \label{fig:figMSPA}
	\end{figure}

Given the good performance of the autoregressive model, some further variations of this concept were investigated. Hence, an extended Autoregressive model was proposed, denoted multi-steps prediction autoregressive (MSPA) model, for which the pre-text task consists in estimating at each timestep $t$ the data from the next timestep $t+1$ until the timestep $t+q$, such that $y^U_t = (\hat{x}^U_{t+1}, \dots, \hat{x}^U_{t+q})$, with $q \in \mathbf{N}^*$, as illustrated in Fig. \ref{fig:figMSPA}.\\

\newpage

Results in Table \ref{tab:mapeEXAR} show that increasing the prediction time horizon in pre-training can improve the predictive performance of the fine-tuned models, when few labelled structures for training are available ($N_{L}^{Train}$ = 5 or 10).
For example, the MSPA model with a time horizon of $q = 30$ and fine-tuned on $N_{L}^{Train} = 5$ structures has an $MAPE$ of about 13\% when $d = 90\%$, while the initial autoregressive model trained under the same conditions has an $MAPE$ of about 22\%. 
Moreover, on very few labelled training structures ($N_{L}^{Train}$ = 5), the results show that the parameter $d$ has a significant influence on the pre-training of the MSPA models: the higher $d$ is, the more the available sequence data is close to the given failure time $T_f$ (see Section 4.1) and the better the RUL estimation performance of the MSPA pre-trained models. This improvement of the MSPA performance compared to the AR one makes sense since when at \textit{d} = 90\%, predicting \textit{q} = 30 timesteps means predicting the strain data until the time of failure. Being able to accurately predict until time of failure facilitates of course the downstream RUL prediction task. However, note that this performance does not hold when more labelled samples for training are available ($N_{L}^{Train}$ greater than 20) by becoming worse than those of the initial autoregressive model and the non-pre-trained model, which does not allow general conclusions to be drawn. One possible explanation for this worsening is that training the DL models with values of \textit{q} greater than 1 is significantly more challenging. Some of the variations seen may then be related to the pre-training phase being not yet fully converged. 

Future work could seek to better control the training convergence of the pre-text task for the MSPA models. As a final remark, note that the authors also tried as outputs of the pre-text task predicting $y^U_t = \hat{x}^U_{t+q}$ only, instead of predicting the entire time-windows ($\hat{x}^U_{t+1}, \dots, \hat{x}^U_{t+q}$). As the results obtained after fine-tuning were similar for the two approaches, in this study only the pre-text task considering the entire time-windows has been presented and described in this paper.\\

\linespread{1}
\begin{table}[!htp]
    \footnotesize

	\setlength{\tabcolsep}{5pt}
	\begin{tabular}{l|llllllll}
	    \hline
		\hline
		 & &  $MAPE$ (\%) & Mean  $\pm$ St. & Dev.\\
		\hline
		\hline
		Labelled structures $N_{L}^{Train}$ & $ 5$ & $ 10$& 20 & 50 &  100\\

		\hline
		\textbf{Pre-trained models (d = 60\%)} & & & \\

		Autoregressive $q$ =  1 &  22.48 $\pm$ 6.06 & 8.08 $\pm$ 1.27  & \textbf{2.63} $\pm$ \textbf{0.80} & $\textbf{1.14} \pm \textbf{0.04}$ &  \textbf{1.03} $\pm$ \textbf{0.06} \\
		MSPA  $q$ =  10& $25.90 \pm 4.02$ &  $8.81 \pm 2.08$  &  $2.86 \pm 0.80$   & $1.22 \pm 0.13$   & $1.04 \pm 0.07$\\
		MSPA  $q$ = 20 & $23.93 \pm 6.54$   &$\textbf{7.81} \pm \textbf{0.56}$ &  $3.55 \pm 0.67$  & $1.34 \pm 0.19$   & $1.06 \pm 0.08$\\
		MSPA  $q$ = 30 & $\textbf{18.88} \pm \textbf{4.02}$   &$8.24 \pm 1.33$ &  $4.95 \pm 0.80$  & $1.57 \pm 0.21$  & $1.23 \pm 0.15$ \\

		\hline
		
		\textbf{Pre-trained models (d = 70\%)} & & &  &  \\

		Autoregressive $q$ =  1 & 24.43 $\pm$ 4.08 & 8.46 $\pm$ 1.53 &  \textbf{2.42} $\pm$ \textbf{0.53}& $\textbf{1.20} \pm \textbf{0.06}$ &  \textbf{1.14 $\pm$ 0.17} \\
		MSPA $q$ = 10 & $21.33 \pm 3.85$ & $8.25 \pm 0.88$ & $4.34 \pm 0.90$  & $1.43 \pm 0.12$  & $1.17 \pm 0.04$ \\
		MSPA $q$ = 20 & $20.99 \pm 5.43$ & $7.95 \pm 1.03$ & $4.62 \pm 0.56$ & $1.70 \pm 0.17$  & $1.22 \pm 0.24$ \\
		MSPA $q$ = 30 & $\textbf{16.56} \pm \textbf{3.16}$ & $\textbf{7.52} \pm \textbf{0.99}$ & $4.79 \pm 0.87$& $2.68 \pm 0.42$ &  $1.45 \pm 0.04$ \\

		\hline
		
		\textbf{Pre-trained models (d = 80\%)} & & & & \\
		
		Autoregressive $q$ =  1 & 26.50 $\pm$ 2.58  &  8.09 $\pm$ 3.28  &  \textbf{2.39} $\pm$ \textbf{0.26} & \textbf{1.39 $\pm$ 0.20} & \textbf{1.07 $\pm$ 0.11} \\
		MSPA $q$ = 10 & $17.75 \pm 4.53$ & $8.96 \pm 0.53$ & $3.53 \pm 1.03$  & $1.47 \pm 0.11$   & $1.18 \pm 0.10$  \\
		MSPA $q$ = 20& $14.84 \pm 3.22$ & $7.03 \pm 1.30$ &  $4.68 \pm 0.77$  & $2.27 \pm 0.41$ & $1.55 \pm 0.17$ \\
		MSPA $q$ = 30 & $\textbf{13.32} \pm \textbf{2.78}$ & $\textbf{5.83} \pm \textbf{0.53}$ &  $4.55 \pm 0.98$ & $2.88 \pm 0.47$ & $2.16 \pm 0.18$ \\

		\hline
		
		\textbf{Pre-trained models (d = 90\%)} & & & &  \\

		Autoregressive $q$ =  1 & 22.69 $\pm$ 2.36 & 8.83 $\pm$ 1.61 &  \textbf{3.39} $\pm$ \textbf{0.40}  & \textbf{1.25 $\pm$ 0.07} & \textbf{0.99 $\pm$ 0.06} \\
		MSPA $q$ = 10& $14.68 \pm 3.55$ & $8.84 \pm 1.73$ & $6.55 \pm 0.85$  & $1.44 \pm 0.05$  & $1.19 \pm 0.06$ \\
		MSPA $q$ = 20 & $13.92 \pm 5.16$ & $\textbf{7.15} \pm \textbf{0.81}$ &  $5.22 \pm 0.59$  & $1.85 \pm 0.25$ & $1.39 \pm 0.14$ \\
		MSPA $q$ = 30 & $\textbf{13.24} \pm \textbf{2.34}$ & $7.45 \pm 0.73$ &  $4.42 \pm 2.04$ & $1.48 \pm 0.21$ & $1.18 \pm 0.09$ \\
		
		\hline

		\textbf{Non-pre-trained models} & & &  & \\

		Autoregressive architecture &  28.09 $\pm$ 1.63   & 21.10 $\pm$ 1.92 &  7.52 $\pm$ 1.59 & 1.15 $\pm$ 0.09 & \textbf{0.79} $\pm$ \textbf{0.09} \\

	\end{tabular}
	\caption{\textit{MAPE} mean values (in \%) plus or minus its standard deviation as a function of the number of labelled structures used and of the training scenario for the autoregressive case. Best performance is represented in bold for each case.}
	\label{tab:mapeEXAR}
\end{table} 
\newpage

\subsubsection{Freezing pre-trained layers during learning}

In the fine-tuning phase of the previous subsection, the weights of the pre-trained layers were frozen, and only the weights of the fine-tuning model were trainable (\textit{i.e. }GRU network for fine-tuning as illustrated in Fig. \ref{fig:SSLAR2}.
Therefore, the authors also sought to investigate the effect of not freezing the pre-trained layers during the fine-tuning phase. Unfreezing the layers means that the pre-text task is basically used to find a good starting point for the training of the full network architecture. As the autoregressive model showed the best performance so far, the investigation was done on this model. In the fine-tuning phase, the model based on the autoregressive structure is composed of 125121 trainable parameters when the pre-trained layers are not frozen, against 25025 trainable parameters when they are frozen. Results in Table \ref{tab:freeze} show that the two approaches perform almost similarly, so it is difficult to determine whether it is better to freeze or not the layers in this RUL estimation problem. Note that both pre-trained models remain better in RUL estimation than their non-pre-trained counterpart, with or without frozen pre-trained layers, which confirms the benefits of the pre-training in all the cases.\\

\begin{figure}[!htp]
		\centering
        \vspace{-0.2cm}
		\includegraphics[width=0.85\textwidth]{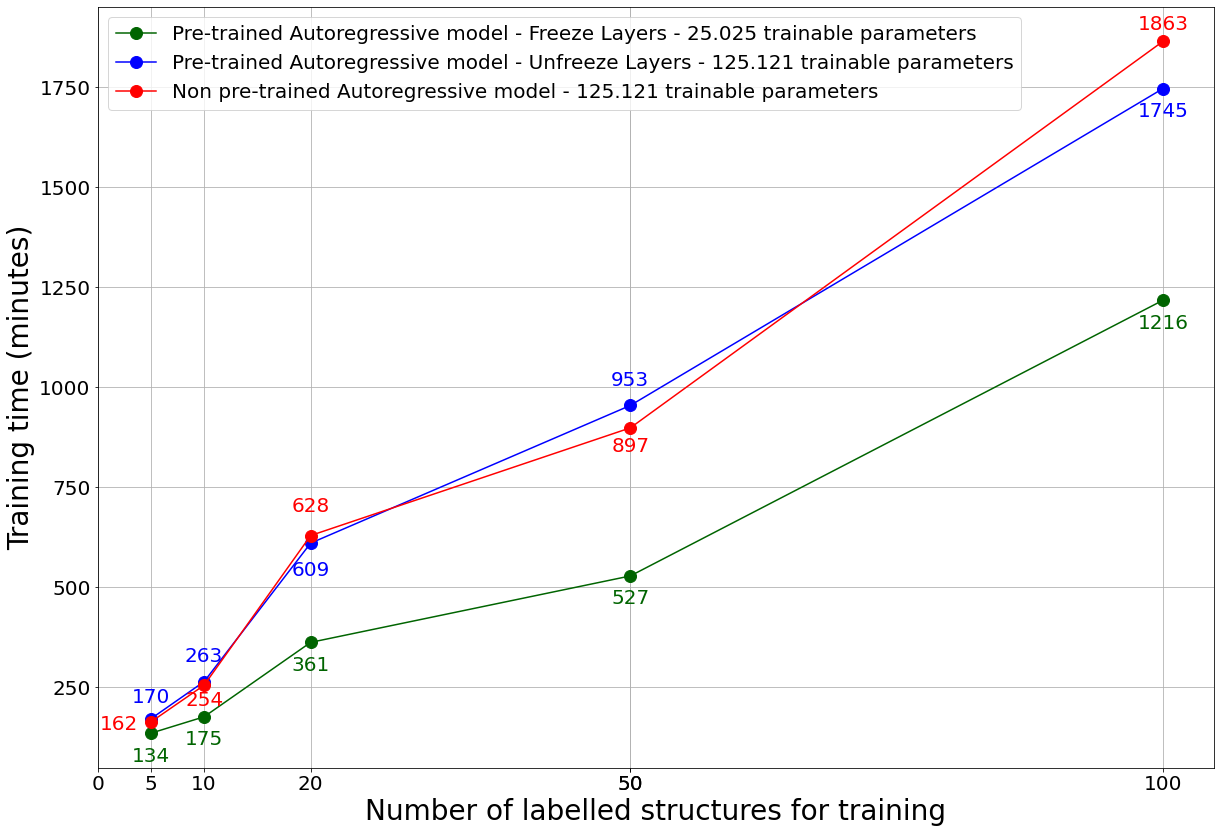}
		\caption{Mean training time during fine-tuning phase (in minutes). }
	   \label{fig:figEX}
	\end{figure}

\linespread{1}
\begin{table}[!htp]
    \footnotesize
    % \hspace{-1.5cm}
	\centering
% 	\setstretch{1}
	\setlength{\tabcolsep}{5pt}
% 	\label{tab:hyperparameters_100_500_1000_structures_TCN}
	\begin{tabular}{l|llllllll}
	    \hline
		\hline
		 & &  $MAPE$ (\%) & Mean  $\pm$ St. & Dev.\\
		\hline
		\hline
		Labelled structures $N_{L}^{Train}$ & $ 5$ & $ 10$& 20 & 50 &  100\\
		\hline

		\textbf{Pre-trained models (d = 60\%)} & & & \\

		Autoregressive - Freeze layers &  22.48 $\pm$ 6.06 & 8.08 $\pm$ 1.27  & \textbf{2.63} $\pm$ \textbf{0.80} & $\textbf{1.14} \pm \textbf{0.04}$ &  \textbf{1.03} $\pm$ \textbf{0.06} \\
		Autoregressive - Unfreeze layers &  \textbf{21.13 $\pm$ 4.87} & \textbf{6.52 $\pm$ 0.56}  & 3.36 $\pm$ 0.80 & $\textbf{1.13} \pm \textbf{0.08}$ &  \textbf{1.05} $\pm$ \textbf{0.06} \\

		\hline
		
		\textbf{Pre-trained models (d = 70\%)} & & &  &  \\
		Autoregressive - Freeze layers& \textbf{24.43 $\pm$ 4.08} & \textbf{8.46 $\pm$ 1.53} &  \textbf{2.42} $\pm$ \textbf{0.53}& $1.20 \pm 0.06$ &  1.14 $\pm$ 0.17 \\
		Autoregressive - Unfreeze layers & 24.75 $\pm$ 3.64 & 8.93 $\pm$ 1.93 &  3.20 $\pm$ 1.18 & $\textbf{1.14} \pm \textbf{0.12}$ &  \textbf{1.01 $\pm$ 0.09} \\

		\hline
		
		\textbf{Pre-trained models (d = 80\%)} & & & & \\
		
		Autoregressive - Freeze layers & \textbf{26.50 $\pm$ 2.58}  &  8.09 $\pm$ 3.28  &  \textbf{2.39} $\pm$ \textbf{0.26} & 1.39 $\pm$ 0.20 & 1.07 $\pm$ 0.11 \\
		Autoregressive - Unfreeze layers & 27.04 $\pm$ 4.74  &  \textbf{7.81 $\pm$ 1.63}  &  2.99 $\pm$ 0.85 & \textbf{1.16 $\pm$ 0.08} & \textbf{1.02 $\pm$ 0.08} \\

		\hline
		
		\textbf{Pre-trained models (d = 90\%)} & & & &  \\

		Autoregressive - Freeze Layers & 22.69 $\pm$ 2.36 & 8.83 $\pm$ 1.61 &  \textbf{3.39} $\pm$ \textbf{0.40}  & $1.25 \pm 0.07$ & 0.99 $\pm$ 0.06 \\
		Autoregressive - Unfreeze Layers & \textbf{22.41 $\pm$ 3.05} & \textbf{8.74 $\pm$ 2.65} &  4.37 $\pm$ 1.29  & $\textbf{1.16} \pm \textbf{0.10}$ & \textbf{0.85 $\pm$ 0.12} \\

		\hline

		\textbf{Non-pre-trained models} & & &  & \\
		Autoregressive architecture &  28.09 $\pm$ 1.63   & 21.10 $\pm$ 1.92 &  7.52 $\pm$ 1.59 & 1.15 $\pm$ 0.09 & \textbf{0.79} $\pm$ \textbf{0.09} \\

	\end{tabular}
	\caption{\textit{MAPE} mean values (in \%) plus or minus its standard deviation as a function of the number of labelled structures used and of the training scenario for the autoregressive case, with or without freezing pre-trained layers. Best performance is represented in bold for each case.}
	\label{tab:freeze}
\end{table} 
% \newpage
Nevertheless, it should be noted that freezing the weights of the pre-trained layers considerably reduces the number of trainable parameters (25025 trainable parameters when the pre-trained layers are frozen, against 125121 trainable parameters when they are not), and therefore reduces the computational complexity during training.
Indeed, Fig. \ref{fig:figEX} shows that freezing the pre-trained layers speeds up the calculations considerably (1.5 to 2 times less time than other models), while both investigated learning approaches in this subsection perform almost similarly as shown in Table \ref{tab:freeze}. \\

% \newpage

\section{Discussion}
\label{sec:discuss}

Based on the previous results we now summarize and discuss some of the effects observed: 

\begin{itemize}
    \item \underline{\textit{Number of unlabelled samples for pre-training:}} Results obtained confirmed that the number of pre-training samples matters. They have shown that pre-training does not always improve predictive performance when the number of pre-training samples is not sufficient, and may even decrease predictive performance (\textit{i.e. negative transfer}). Nevertheless, these investigations indicated that as the number of unlabelled samples increases, the pre-training becomes more efficient and allows to have better results than a non-pre-trained model when few labelled structures are available. A research direction to further improve the pre-training process would be to select the available unlabelled sample, with the aim of extracting the most useful features from the data and avoiding over-fitting, in the spirit of deep active learning \cite{ren2021survey}. On the application we considered, for example, it could be interesting to implement an adaptive pre-training strategy to select the training samples and remove unnecessary samples (\textit{e.g.} sensor data with very little variation).

    \item \textit{\underline{Pre-text task:}} In this work, several pre-training tasks were compared (\textit{i.e.} input signal estimation and prediction of the future outcome of a sequence) in order to identify which one is most appropriate for the considered case study, and by extension for other engineering case studies using time series sensor data. Experiments have shown that autoregressive pre-training tasks outperform the AE model in pre-training, and capture useful representations from the sensor data (\textit{i.e.} temporal dependencies of the input signal) for RUL estimation tasks. Moreover, results showed that increasing the prediction time horizon of autoregressive models in pre-training can improve the predictive performance, notably when few labelled structures are available. In next steps, it would be interesting to explore other pre-text tasks (\textit{e.g.} Contrastive learning, which aims at learning similar or dissimilar representations from source data \cite{jaiswal_survey_2021}). Another interesting research direction would be to embed a Bayesian framework to the models in the pre-training phase (\textit{e.g.} Variational Autoencoders \cite{kingma2013auto}), in order to address the random nature of the data such as noise or measurement errors (\textit{i.e.} aleatoric uncertainty). Note that the VAE model, although it has shown promising results in learning meaningful representations from raw unlabelled data \cite{zhu2020s3vae,dhariwal2020jukebox}, has been studied and implemented in this work but the results were unsatisfactory. This could be due to the stochastic nature of the model during sampling: thus it requires further investigation in the future.
    
\end{itemize}

\section{Conclusion}
\label{sec:conclusions_outlook}

% In this paper, a Self-Supervised Learning approach for fatigue damage prognostics problem was proposed and investigated. The approach is based on combination of a pretext and a fine-tuning task. In the pretext task a model is trained using a large number of raw (unlabelled) sensor data. Then, in the fine tuning task, a new model is adjusted based on a smaller number of data, labelled with corresponding RUL. Multiple scenarios were investigated within this framework, including varying the pretext task, the models and the dataset properties. \\

In this paper, a Self-Supervised Learning approach for fatigue damage prognostics problem was proposed and investigated. The approach is based on combination of a pretext and a downstream task. In the pretext task a model is trained using a large number of raw (unlabelled) sensor data with the aim of learning general representations linked to the degradation process. No RUL data is available during this pretext task, only raw sensor data (strains in our case), as the data is obtained only on structures that have not yet reached failure. Then, in a subsequent downstream task, a new model, aimed at predicting the RUL, is adjusted based on the previously pretrained model and based on a limited number of labelled RUL data obtained on structures that have reached failure. Multiple scenarios were investigated within this framework, including varying the pretext task, the models and the dataset properties.\\

The results obtained showed that self supervised learning is efficient in prognostics and can improve RUL estimation performances especially when only a limited amount of labelled data is available. Overall, these investigations indicate that pre-trained models are able to significantly outperform the respective non-pre-trained counterpart models in the RUL prediction task, while at the same time lowering training computational costs. Accordingly, the proposed approach can significantly reduce the need for labelled data for a given prediction accuracy, or alternatively significantly improve the prediction accuracy for the same amount of (limited) labelled data. Furthermore, the authors of this paper believe that the potential of this learning approach will benefit researchers in a variety of similar engineering fields using sensors or time series data (\textit{e.g.} in energy \cite{jain2006exploiting}, or water and environmental engineering \cite{borzooei2019data}) and that it can be reused to overcome the lack of available labelled data.\\

In next steps, it would be interesting to explore other pre-text tasks (\textit{e.g. }contrastive learning, ensemble learning, etc.) or other models (\textit{e.g. }masked autoencoders), adaptive activation functions \cite{jagtap2020adaptive, jagtap2020locally, jagtap2022deep, jagtap2022important}). Furthermore, as uncertainty quantification remains a challenging and ubiquitous task in real-world ML applications (\textit{e.g.} in engineering domains such as transportation engineering \cite{mazloumi2011prediction} or water and environmental applications \cite{ghiasi2022uncertainty}), it could be interesting to use Bayesian machine learning models in SSL (\textit{e.g.} Deep Gaussian Processes) to quantify uncertainty in downstream prognostics tasks. Another future work perspective consists in combining strain data with other type of sensor data (\textit{e.g.} ultrasound mappings) in a self-supervised framework in order to further improve prediction results. Future work is also aimed at investigating how the proposed self-supervised prognostics framework behaves on an actual engineering problem involving real-world data.

\section*{Acknowledgements}

This work was partially funded by the French ``Occitanie Region" under the Predict project. This funding is gratefully acknowledged. This work has been partly carried out on the supercomputers PANDO (ISAE-SUPAERO, Toulouse) and Olympe (CALMIP, Toulouse, project n°21042). Authors are grateful to ISAE-SUPAERO and CALMIP for the hours allocated to this project.

% \section{}
% \label{}

\appendix

\section{Deep Gated Recurred Unit networks}
\label{GRU}

\begin{figure}[!htp]
\centering
\includegraphics[scale=1.]{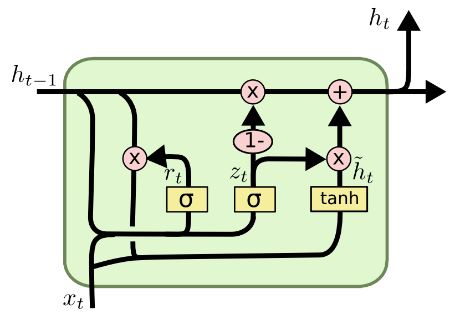}
\caption{GRUs architecture. Source in \cite{colah}}

    \label{fig:gru}
\end{figure}

Introduced by Cho et al. \cite{cho2014learning} Gated Recurrent Unit, or GRU, are a variant of recurrent neural networks which solves the time-delay problem existing in traditional RNNs. This approach has gained in popularity in recent years due to its relative simplicity (\textit{i.e.} lower complexity and faster computation \cite{rana2016gated}), while the same ability to capture the mapping relationships among time series data  \cite{yamak2019comparison}. The structure of the GRU network is shown in Fig. \ref{fig:gru}. \\

The formulas that govern the computation happening in a GRU network are as follow\cite{cho2014learning}:

\begin{equation}
\begin{split}
\begin{aligned}
z_t &= \sigma(W_z X_t + U_z h_{t-1})  \\
r_t &= \sigma(W_f X_t + U_f h_{t-1}) \\
\Tilde{h}_t &= \text{\textit{tanh}}(W_{\Tilde{h}} X_t + U_{\Tilde{h}} [h_{t-1}*r_t])\\
h_t &= z_t * \Tilde{h}_{t} + (1-z_t)*h_{t-1} 
\end{aligned}
\end{split}
\end{equation}

where $x_t$ is the input sequence at time step $t$, $h_t$ a hidden state, $z_t$ the update gate, $r_t$ the reset gate, $\Tilde{h}_t$ a cell state,  $\sigma$(.) represents the sigmoid activation function and \textit{tanh}(.) the hyperbolic tangent non-linear function, $W$ and $U$ denote the weight matrices which are learned during training. The pink circles represent pointwise operations (\textit{e.g.} addition, multiplication). The idea behind the GRU network is that in each unit, the update gate $z_t$ must select whether the hidden state $h_t$ is to be updated with
a new hidden state $\Tilde{h}_t$; the reset gate $r_t$ must decide
whether the previous hidden state $h_{t-1}$ is ignored. More details can be found in \cite{cho2014learning}.

 \bibliographystyle{elsarticle-num} 
 \bibliography{ref}
% \biboptions{sort&compress}
\biboptions{compress}





% \end{thebibliography}
\end{document}

%% file: acronyms.tex
\newacronym{nlp}{NLP}{Natural Language Processing}
\newacronym{phm}{PHM}{Prognostics and Health Management}
\newacronym{rul}{RUL}{Remaining Useful Life}
\newacronym{rnn}{RNN}{Recurrent Neural Network}
\newacronym{lstm}{LSTM}{Long short-term memory}
\newacronym{gru}{GRU}{Gated Recurrent Unit}
\newacronym{cnn}{CNN}{Convolutional Neural Network}
\newacronym{tcn}{TCN}{Temporal Convolutional Network}
\newacronym{dl}{DL}{Deep Learning}
\newacronym{erm}{ERM}{Elman Recurrent Network}
\newacronym{mlp}{MLP}{Multilayer Perceptron}
\newacronym{lrdecay}{lrDecay}{Learning rate decay}
\newacronym{eol}{EOL}{End of life}